\documentclass[10pt,journal,compsoc]{IEEEtran}
\ifCLASSOPTIONcompsoc
  \usepackage[nocompress]{cite}
\else
  \usepackage{cite}
\fi
\ifCLASSINFOpdf
   \usepackage[pdftex]{graphicx}
   \graphicspath{{../pdf/}{../jpeg/}}
   \DeclareGraphicsExtensions{.pdf,.jpeg,.png}
\else
\fi
\usepackage{amsmath}
\usepackage{color,xcolor}
\usepackage{subfigure}

\usepackage{mathtools}
\usepackage{amssymb}

\ifCLASSOPTIONcompsoc
 \usepackage[caption=false,font=footnotesize,labelfont=sf,textfont=sf]{subfig}
\else
 \usepackage[caption=false,font=footnotesize]{subfig}
\fi

\usepackage{multirow}
\begin{document}
%
\title{RankSRGAN: Super Resolution Generative Adversarial Networks with Learning to Rank}

%
%
%
%

\author{Wenlong Zhang,
        Yihao Liu,
        Chao Dong,
        Yu Qiao
\IEEEcompsocitemizethanks{\IEEEcompsocthanksitem Wenlong Zhang, Yihao Liu, Chao Dong and Yu Qiao are with ShenZhen Key Lab of Computer Vision and Pattern Recognition, SIAT-SenseTime Joint Lab,Shenzhen Institutes of Advanced Technology, Chinese Academy of Sciences, China. 

\IEEEcompsocthanksitem Yu Qiao is with Shanghai AI Lab, Shanghai, China.

\IEEEcompsocthanksitem E-mail: \{wl.zhang1, yh.liu4, chao.dong, yu.qiao\}@siat.ac.cn \protect\\
Corresponding author: Chao Dong (chao.dong@siat.ac.cn).

}

}

%
%

\markboth{Journal of \LaTeX\ Class Files}%
{Shell \MakeLowercase{\textit{et al.}}: Bare Demo of IEEEtran.cls for Computer Society Journals}
%
\ifCLASSOPTIONpeerreview
\begin{center} \bfseries EDICS Category: 3-BBND \end{center}
\fi



\IEEEtitleabstractindextext{%
\begin{abstract}
Generative Adversarial Networks (GAN) have demonstrated the potential to recover realistic details for single image super-resolution (SISR). To further improve the visual quality of super-resolved results, PIRM2018-SR Challenge employed perceptual metrics to assess the perceptual
quality, such as PI, NIQE, and Ma. However, existing methods cannot directly optimize these indifferentiable perceptual metrics, which are shown to be highly correlated with
human ratings.  To address the problem, we propose Super-Resolution Generative Adversarial Networks with Ranker (RankSRGAN) to optimize generator in the direction of different perceptual metrics. Specifically, we first train a Ranker which can learn the behaviour of perceptual metrics and then introduce a novel rank-content loss to optimize the perceptual quality. The most appealing part is that the proposed method can combine the strengths of different SR methods to generate better results. Furthermore, {we extend our method to multiple Rankers} to provide multi-dimension constraints for the generator. Extensive experiments show that RankSRGAN achieves visually pleasing results and reaches state-of-the-art performance in perceptual metrics and quality. 
\textcolor{magenta}{Project page: https://wenlongzhang0517.github.io/Projects/RankSRGAN}
\end{abstract}

}

\maketitle

\IEEEdisplaynontitleabstractindextext

\ifCLASSOPTIONpeerreview
\begin{center} \bfseries EDICS Category: 3-BBND \end{center}
\fi
%
\IEEEpeerreviewmaketitle

\IEEEraisesectionheading{\section{Introduction}\label{sec:introduction}}

%
%
%
%

\IEEEPARstart{S}{ingle} image super resolution aims at reconstructing/generating a high-resolution (HR) image from a low-resolution (LR) observation. Thanks to the strong learning capability, Convolutional Neural Networks (CNNs) have demonstrated superior performance   \cite{dong2014learning,lim2017enhanced,zhang2018residual,haris2018deep,zhang2018rcan,wang2018recovering,Wang2018ESRGAN,guo2020closedloop} to the conventional example-based \cite{zhang2012single,Freeman2011Image,Freeman2002Example,1315043,5459271} and interpolation-based \cite{zhang2006edge,6161647,6414620} algorithms. Recent CNN-based methods can be divided into two groups. The first one regards SR as a reconstruction problem and adopts mean squared error (MSE) as the loss function to achieve high PSNR values. However, due to the conflict between the reconstruction accuracy and visual quality, they tend to produce overly smoothed/sharpened images. To favour better visual quality, the second group casts SR as an image generation problem \cite{ledig2017photo}. By incorporating the perceptual loss \cite{bruna2015super,johnson2016perceptual} and adversarial learning \cite{ledig2017photo}, these perceptual SR methods have the potential to generate realistic textures and details, thus attracted increasing attention in recent years. 

\begin{figure}[t]
\setlength{\abovecaptionskip}{-0.3cm}
\setlength{\belowcaptionskip}{-0.2cm}
\begin{center}
	\includegraphics[width=1\linewidth]{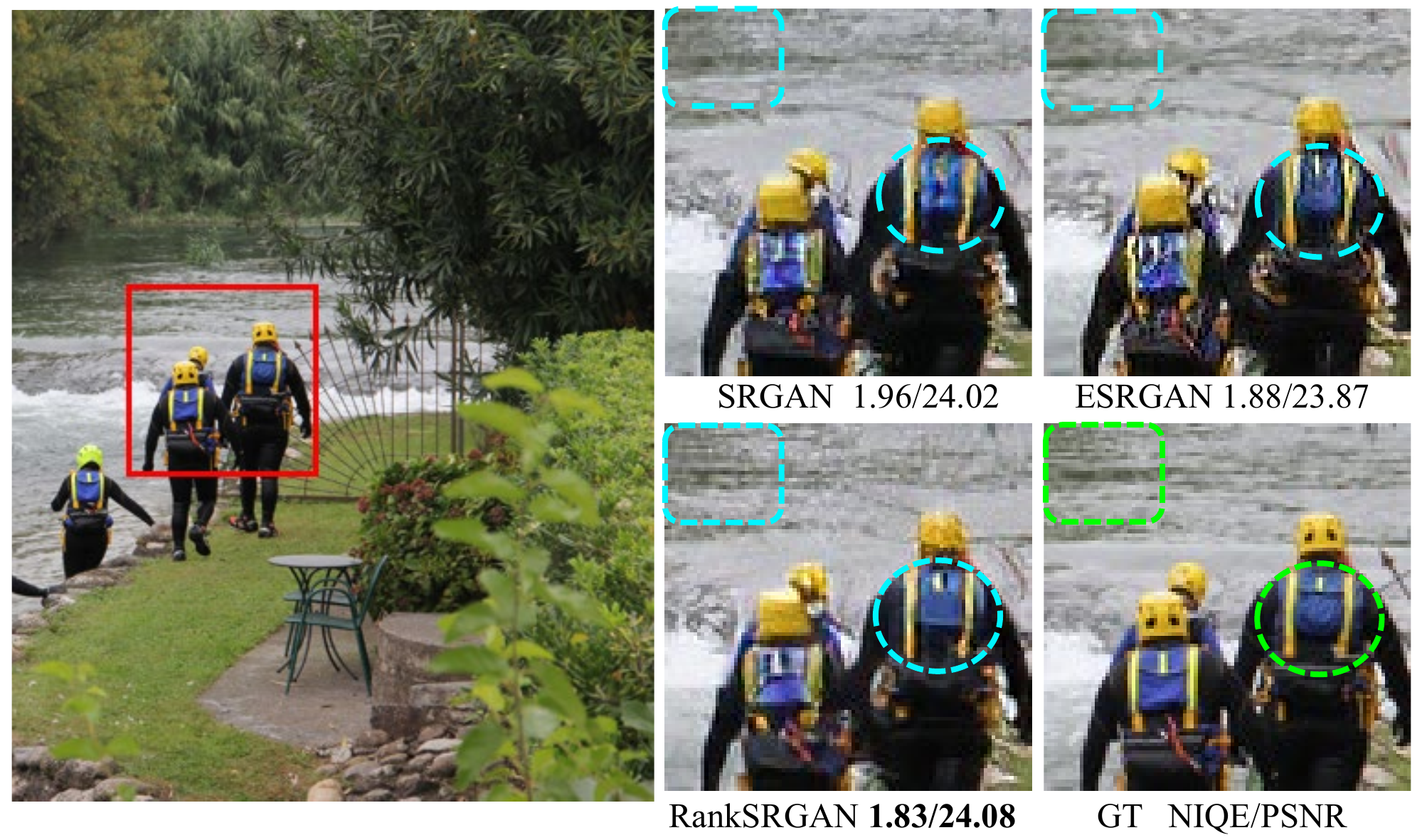} 

\end{center}
   \caption{The comparison of RankSRGAN and the state-of-the-art perceptual SR methods on $\times$4. NIQE: lower is better. PSNR: higher is better.}
\label{fig:1}
\label{fig:onecol}

\end{figure}

The most challenging problem faced with perceptual SR methods is the evaluation. Most related works resort to user study for subjectively evaluating the visual quality \cite{blau20182018, wang2018recovering}. However, without an objective metric like PSNR/SSIM, it is hard to compare different algorithms on a fair platform, which largely prevents them from rapid development. To address this issue, a number of no-reference image quality assessment (NR-IQA) metrics are proposed, and some of them are proven to be highly correlated with human ratings \cite{blau20182018}, such as NIQE \cite{mittal2013making} (correlation 0.76) and PI \cite{blau20182018} (correlation 0.83). Specially, the PIRM2018-SR challenge \cite{blau20182018} introduced the PI metric as perceptual criteria and successfully ranked the entries. Nevertheless, most of these NR-IQA metrics are not differentiable (e.g., they include hand-crafted feature extraction or statistic regression operation), making them infeasible to serve as loss functions. Without considering NR-IQA metrics in optimization, existing perceptual SR methods could not show stable performance in the orientation of objective perceptual criteria. 

To overcome this obstacle, we propose a general and differentiable model -- Ranker, which could mimic any perceptual characteristic and provide a clear goal (as loss function) for optimizing perceptual quality. Specifically, Ranker is a Siamese CNN that simulates the behaviour of the perceptual characteristic by learning to rank approach \cite{burges2005learning}. For example, we could  employ Ranker to learn the ranking orders of perceptual metrics instead of the absolute values. Just like in the real world, people tend to rank the quality of images rather than give a specific value. We equip Ranker with the standard SRGAN \cite{ledig2017photo} model and form a new perceptual SR framework -- RankSRGAN (Super-Resolution Generative Adversarial Networks with Ranker). In addition to SRGAN, the proposed framework has a rank-content loss using a well-trained Ranker to measure the output image quality. Then the SR model can be stably optimized in the orientation of specific perceptual metrics.

To train the proposed Ranker, we prepare another training dataset with rank label. Then the Ranker, with a Siamese-like architecture, could learn these ranking orders with high accuracy. The effectiveness of the Ranker is largely determined by the selected rank dataset. To achieve the best perceptual performance, we adopt two state-of-the-art perceptual SR models -- SRGAN \cite{ledig2017photo} and ESRGAN \cite{Wang_2018_ECCV_Workshops}. As the champion of PIRM2018-SR challenge \cite{blau20182018}, ESRGAN is superior to SRGAN on average scores, but could not outperform SRGAN on all test images. When evaluating with NIQE \cite{mittal2013making}, we obtain mixed orders for these two methods. Then the Ranker will favour different algorithms on different images, rather than simply classifying an image into a binary class (SRGAN/ESRGAN). After adopting the rank-content loss, the generative network will output results with higher ranking scores. In other words, the learned SR model could combine the better parts of SRGAN and ESRGAN, and achieve superior performance both in perceptual metric and visual quality. Figure \ref{fig:1} shows an example of RankSRGAN, which fuses the imagery effects of SRGAN and ESRGAN and obtains better NIQE score. 
We have done comprehensive ablation studies to further validate the effectiveness of the proposed method. First, we distinguish our Ranker from the regression/classification network that could also mimic the perceptual metric. Then, we train and test RankSRGAN with several perceptual metrics (i.e. NIQE \cite{mittal2013making}, Ma \cite{ma2017learning}, PI \cite{blau20182018}). We further show that adopting different SR algorithms to build the dataset achieves different performance. Besides, we have also investigated the effect of different loss designs and combinations. With proper formulation, our method can clearly surpass ESRGAN and achieve state-of-the-art performance.

The short version of this work has been published in ICCV2019 \cite{Zhang_2019_ICCV}. We have made significant changes to that conference paper. First, in the main method section, we add detailed analysis on the working mechanism of the proposed Ranker, discriminator and perceptual loss. Then we extend the rank dataset to a more general form, which does NOT rely on objective metrics or SR algorithms. Specifically, we propose two new formulations, one with image interpolation and the other with artificial distortions. With different rank datasets, the Ranker could provide different constraints on SRGAN, and produce images with various perceptual characteristics. To further take advantage of different Rankers, we have also proposed RankSRGAN in multiple dimensions. This is achieved by {constructing multiple Rankers with different rank datasets.} Experiments show that the multi-dimension RankSRGAN could favour multiple metrics simultaneously and combine the imagery effects of different models. All these extensions can largely improve the flexibility and generalization ability of RankSRGAN. 

In summary, the contributions of this paper are four-fold.

(1) We propose a general perceptual SR framework -- RankSRGAN that could optimize generator in the direction of indifferentiable perceptual metrics and achieve the state-of-the-art performance. 

(2) We, for the first time, utilize results of other SR methods to build the training dataset. The proposed RankSRGAN could combine the strengths of different SR methods and generate better results. 

(3)The proposed SR framework is highly flexible and can produce diverse results given different rank datasets, perceptual metrics, and loss combinations

(4) We propose a {simple and effective strategy with multiple Rankers} to provide multi-dimension constraints for the generator.

\section{Related work}
\label{sec:2}
\textbf{Super resolution.} Since Dong et al. \cite{dong2014learning} first introduced convolutional neural networks (CNNs) to the SR task, a series of learning-based works \cite{tong2017image,zhang2012single,haris2018deep,kim2016accurate, He_2019_CVPR, feng2019suppressing, gu2019blind} have achieved great improvements in terms of PSNR. For example, Kim et al. \cite{kim2016accurate} propose a very deep network VDSR with gradient clipping. The residual and dense block \cite{lim2017enhanced,zhang2018residual} are explored to improve the super-resolved results. Furthermore, Channel attention \cite{zhang2018rcan} also be introduced to SR problem and achieve state-of-the-art in terms of PSNR. These methods focus on the design of network architecture, training strategy and effective
feature utilization. However, the super-resolved results of these methods appear overly-smooth due to the MSE loss. To solve this problem, SRGAN \cite{ledig2017photo} is proposed to generate more realistic images. Then, texture matching  \cite{sajjadi2017enhancenet} and semantic prior \cite{wang2018recovering} are introduced to improve perceptual quality. Furthermore, the perceptual index  \cite{blau20182018} consisting of NIQE \cite{mittal2013making} and  Ma \cite{ma2017learning} is adopted to measure the perceptual SR methods in the PIRM2018-SR Challenge at ECCV \cite{blau20182018}. In the Challenge  \cite{blau20182018}, ESRGAN \cite{Wang_2018_ECCV_Workshops} achieves the state-of-the-art performance by improving network architecture,  relativistic average discriminator and perceptual loss. However, the relativistic average discriminator results in the unstable training procedure although relativistic discriminator can achieve more better quantitative result than vanilla discriminator in term of perceptual index. Rad et al. \cite{Rad_2019_ICCV} proposed a decoder to construct a targeted perceptual loss to guide the generator in a semantic orientation, which includes object, background and boundary. Zhou et al.\cite{Zhou_2019_ICCV} proposed a kernel modelling super-resolution network to improve generalization and robustness of deep super resolution CNNs on a real scenario. Ma et al.\cite{ma2020structure} utilized a gradient loss to help generative networks to concentrate geometric structures.

\textbf{CNN for NR-IQA.} Most of the No-reference Image Quality Assessment (NR-IQA) can be implemented by learning-based models, which extract hand-crafted features from Natural Scene Statistics (NSS), such as CBIQ \cite{ye2011no}, NIQE \cite{mittal2013making}, and Ma \cite{ma2017learning}, etc. In  \cite{li2011blind}, Li et al. develop a general regression neural network to fit human subjective opinion scores with pre-extracted features. Kang et al.  \cite{kang2014convolutional,bosse2018deep} integrate a general CNN framework which can predict image quality on local regions. In addition, Liu et al.  \cite{liu2017rankiqa} propose RankIQA to tackle the problem of lacking human-annotated data in NR-IQA. They first generate large distorted images in different distortion level. Then they train a Siamese Network to learn the rank of the quality of those images, which can improve the performance of the image quality scores. For SR task, we expect to introduce the rank information of perceptual metrics to SR model training.

\textbf{Learning to rank.} It has been demonstrated that learning to rank approach is effective in information retrieval, document searching and image matching \cite{liu2009learning, cao2006adapting, freund2003efficient, herbrich1999support, burges2005learning}. In \cite{burges2005learning},
Burges et al. extended the work of Ranking SVM \cite{herbrich1999support} and
proposed a new model named RankNet. They employed
Neural Networks as classifier and simply adopted Cross Entropy
as loss function to train the model. Furthermore, the
idea of learning to rank can be also applied in computer
vision. For instance, Devi Parikh et al.  \cite{parikh2011relative} model relative attributes using a well-learned ranking function. Yang et al. \cite{yang2016deep} first employ CNN for relative attribute ranking in a unified framework. One of the most relevant studies to our work is RankCGAN \cite{saquil2018ranking}, which investigates the use of GAN to tackle the task of image generation with semantic attributes. Unlike standard GANs that generate the image from noise input (CGAN \cite{mirza2014conditional}), RankCGAN incorporates a pairwise Ranker into CGAN architecture so that it can handle continuous attribute values with subjective measures. 

Inspired by those works, we propose a Ranker (R) to learn the behaviour of perceptual metrics. In particular, the Ranker can guide SR network to optimize in the orientation of perceptual metrics. Please see the details in Section \ref{sec:3}.
\begin{figure*}[ht!]
\setlength{\abovecaptionskip}{-0.4cm}
\setlength{\belowcaptionskip}{-0.2cm}
\begin{center}
\includegraphics[width=1\linewidth]{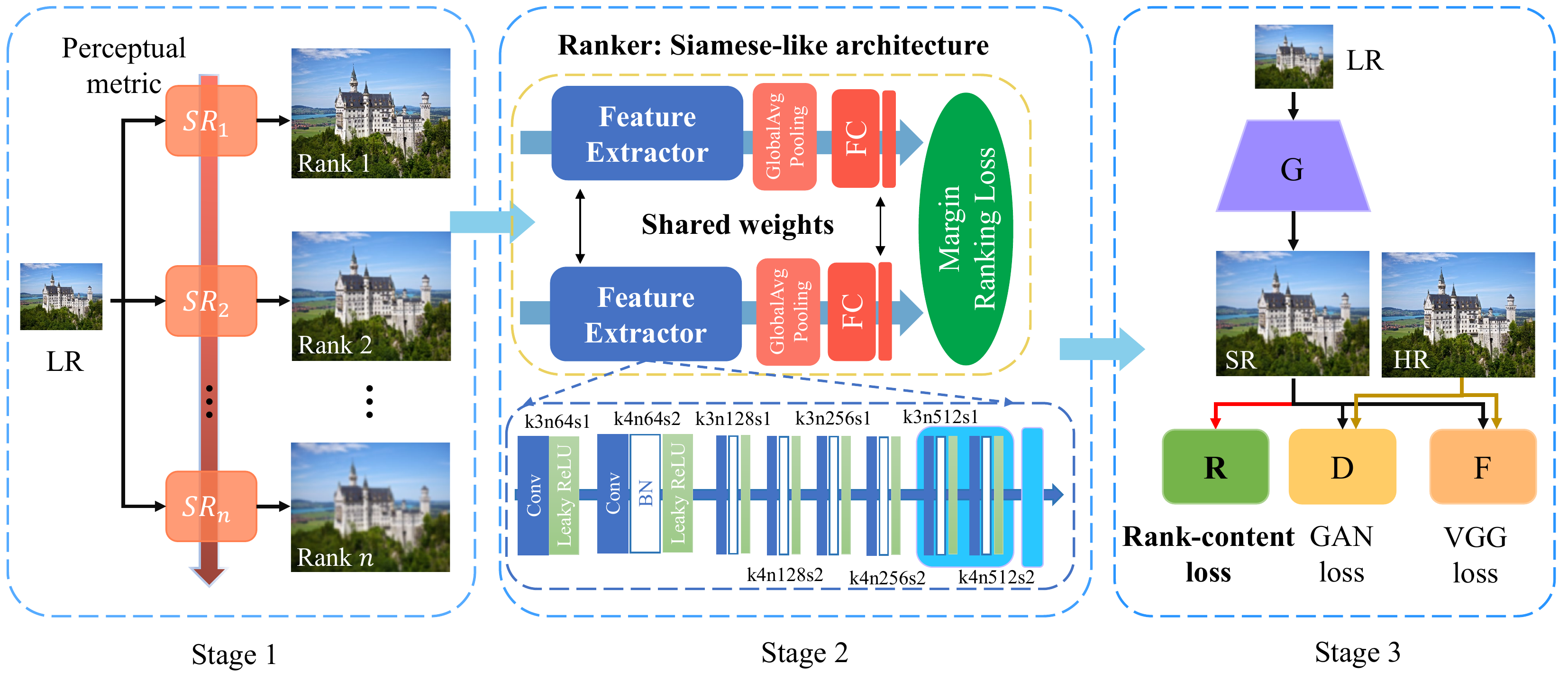} 
\end{center}
   \caption{Overview of the proposed method. \textbf{Stage 1}: Generate pair-wise rank images by different SR models in the orientation of perceptual metrics. \textbf{Stage 2}: Train Siamese-like Ranker network. \textbf{Stage 3}: Introduce rank-content loss derived from well-trained Ranker to guide GAN training. RankSRGAN consists of a generator(G), discriminator(D), a fixed Feature extractor(F) and Ranker(R). }
\label{fig:2}

\end{figure*}

\section{Method}
\label{sec:3}


\subsection{Overview of RankSRGAN}
The proposed framework is built upon the GAN-based \cite{ledig2017photo} SR approach, which consists of a generator and a discriminator. The discriminator network tries to distinguish the ground-truth images from the super-resolved results, while the generator network is trained to fool the discriminator. To obtain more natural textures, we propose to add additional constraints on the standard SRGAN \cite{ledig2017photo} by exploiting the prior knowledge of perceptual metrics to improve the visual quality of output images. The overall framework of our approach is depicted in Figure \ref{fig:2}. The pipeline involves the following three stages:

\textbf{Stage 1: Generate pair-wise rank images.} First, we employ different SR methods to generate super-resolved images on public SR datasets. Then we apply a chosen perceptual metric (e.g. NIQE) on the generated images. After that, we can pick up two images of the same content to form a pair and rank the pair-wise images according to the quality score calculated by the perceptual metric. Finally, we obtain the pair-wise images and the associated ranking labels. More details will be presented in Section \ref{4.1}.

\textbf{Stage 2: Train Ranker.} The Ranker adopts a Siamese architecture to learn the beheviour of perceptual metrics and the network structure is depicted in Section \ref{section:3.2}. We adopt margin-ranking loss, which is commonly used in ``learning to rank'' \cite{burges2005learning}, as the cost function to optimize Ranker. The learned Ranker is supposed to have the ability to rank images according to their perceptual scores.

\textbf{Stage 3: Introduce rank-content loss.} Once the Ranker is well-trained, we use it to define a rank-content loss for a standard SRGAN to generate visually pleasing images. Please see the rank-content loss in Section \ref{3.3}.

\subsection{Ranker}
\label{section:3.2}
\textbf{Rank dataset.} Similar to \cite{choi2018deep,liu2017rankiqa}, we use super-resolution results of different SR methods to represent different perceptual levels.
With a given perceptual metric, we can rank these results in a pair-wise manner. Picking any two SR images, we can get their ranking order according to the quality score measured by the perpetual metric. These pair-wise data with ranking labels form a new dataset, which is defined as the rank dataset. Then we let the proposed Ranker learn the ranking orders. Specifically, given two input images $y_{1}$, $y_{2}$, the ranking scores $s_1$ and $s_2$ can be obtained by

\begin{small}
\begin{equation}
{s_1} = R({y_1};{\Theta _R})
\end{equation}
\begin{equation}
{s_2} = R({y_2};{\Theta _R}),
\end{equation}
\end{small}where ${\Theta _R}$ represents the network weights and $R(.)$ indicates the mapping function of Ranker. In order to make the Ranker output similar ranking orders as the perceptual metric, we can formulate:
\begin{small}
\begin{equation}
\left\{
             \begin{array}{lr}
             s_1<s_2  & if \quad m_{y_1}<m_{y_2}\\
             s_1>s_2  & if \quad m_{y_1}>m_{y_2}\\ 
             \end{array}
\right.,
\end{equation}
\end{small}where $ m_{y_1} $ and $ m_{y_2} $ represent the quality scores of image $y_1$ and image $y_2$, respectively. A well-trained Ranker could guide the SR model to be optimized in the orientation of the given perceptual metric.

\textbf{Siamese architecture.} The Ranker uses a Siamese-like architecture \cite{bromley1994signature,chopra2005learning,zagoruyko2015learning}, which is effective for pair-wise inputs. {It is designed to simulate the behavior of indifferentiable perceptual metrics by learning to rank approach.} The architecture of Ranker is shown in Figure \ref{fig:2}. It has two identical network branches which contain a series of convolutional, LeakyReLU, pooling and fully-connected layers. Here we use a Global Average Pooling layer after the Feature Extractor, thus the architecture can get rid of the limit of input size. To obtain the ranking scores, we employ a fully-connected layer as a regressor to quantify the rank results. Note that we do not aim to predict the real values of the perceptual metric since we only care about the ranking information.  Finally, the outputs of two branches are passed to the margin-ranking loss module, where we can compute the gradients and apply back-propagation to update parameters of the whole network.

\textbf{Optimization.} To train Ranker, we {only need to} employ margin-ranking loss that is commonly used in sorting problems \cite{yang2016deep,liu2017rankiqa}. The margin-ranking loss is given below:
\begin{small}
\begin{equation}
\setlength{\abovedisplayskip}{3pt} 
\setlength{\belowdisplayskip}{3pt}
\begin{split}
L({s_1},&{s_2},\gamma) = \max (0,({s_1} - {s_2})*\gamma + \varepsilon )\\
&\left\{
             \begin{array}{lr}
             \gamma\,\,=-1  & if \quad m_{y_1}>m_{y_2}\\
             \gamma\,\,=1  & if \quad m_{y_1}<m_{y_2}\\ 
             \end{array}
\right.,
\end{split}
\end{equation}
\end{small}where the $s_1$ and $s_2$ represent the ranking scores of pair-wise images. The $\gamma$ is the rank label of the pair-wise training images. The margin $\varepsilon$ can control the distance between $s_1$ and $s_2$. Therefore, the $N$ pair-wise training images can be optimized by:

\begin{small}
\begin{equation}
\setlength{\abovedisplayskip}{-9pt} 
\setlength{\belowdisplayskip}{-5pt}
\begin{split}
\hat{\Theta}=&\mathop{\arg\min}_{\Theta_R}\frac{1}{N}\sum_{i=1}^N L(s_1^{(i)},s_2^{(i)};\gamma^{(i)})\\
=&\mathop{\arg\min}_{\Theta_R}\frac{1}{N}\sum_{i=1}^NL(R({y_1^{(i)}};{\Theta _R}),R({y_2^{(i)}};{\Theta _R});\gamma^{(i)})
\end{split}
\end{equation}
\end{small}

\begin{figure}[t]
\setlength{\abovecaptionskip}{-0.1cm}
\setlength{\belowcaptionskip}{-0.6cm}
\begin{center}
	\includegraphics[width=1\linewidth]{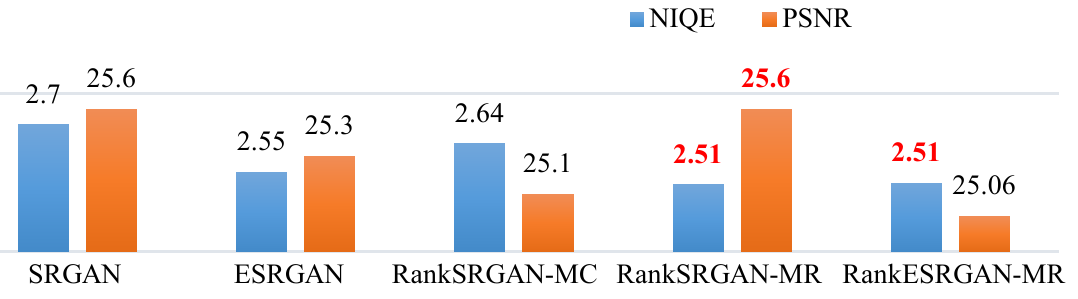} 

\end{center}
   \caption{The NIQE of RankSRGAN-MR exceeds that of SRGAN, ESRGAN and RankSRGAN-MC.}
\label{fig:5}

\end{figure}

\subsection{RankSRGAN}
\label{3.3}
RankSRGAN consists of a standard SRGAN and the proposed Ranker, as shown in Figure \ref{fig:2}. Compared with existing SRGAN, our framework simply adds a well-trained Ranker to constrain the generator in SR space. To obtain visually pleasing super-resolved results, adversarial learning \cite{ledig2017photo,sajjadi2017enhancenet} is applied to our framework where the generator and discriminator are jointly optimized with the objective given below:
\begin{small}
\begin{equation}
\setlength{\abovedisplayskip}{3pt} 
\setlength{\belowdisplayskip}{3pt}
\mathop{\min}_{\theta}\mathop{\max}_{\eta} E_{y\backsim p_{HR}}log{D_{\eta}}(y)+E_{y\backsim p_{LR}}log(1-D_\eta(G_\theta(x))),
\end{equation}
\end{small}where $ p_{HR} $ and $ p_{LR} $ represent the probability distributions of $ HR $ and $ LR $ samples, respectively. In order to demonstrate the effectiveness of the proposed Ranker, we do not use complex architectural designs of GAN but use the general SRGAN \cite{ledig2017photo}.

\textbf{Perceptual loss.} In \cite{dosovitskiy2016generating,johnson2016perceptual}, the perceptual loss is proposed to measure the perceptual similarity between two images. Instead of computing distances in image pixel space, the images are first mapped into feature space and the perceptual loss can be presented as:
\begin{small}
\begin{equation}
\setlength{\abovedisplayskip}{3pt} 
\setlength{\belowdisplayskip}{1pt}
L_P=\sum_i||\phi(\hat{y_i})-\phi(y_i)||_2^2,
\end{equation}
\end{small}where $\phi(y_i)$ and $\phi(\hat{y_i})$ represent the feature maps of HR and SR images, respectively. The $\phi$ can be obtained by the 5-th convolution (before maxpooling) layer within VGG19 network \cite{simonyan2014very}.

\textbf{Adversarial loss.} Adversarial training \cite{ledig2017photo,sajjadi2017enhancenet} is recently used to produce natural-looking images. A discriminator is trained to distinguish the real image from the generated image. This is a minimax game approach where the generator loss $ L_G $ is defined based on the output of discriminator:
\begin{small}
\begin{equation}
\setlength{\abovedisplayskip}{4pt} 
\setlength{\belowdisplayskip}{4pt}
L_G=-logD(G(x_i)),
\end{equation}
\end{small}where $ x_i $ is the $ LR $ image, $ D(G(x_i)) $ represents the probability of the discriminator over all training samples.

\textbf{Rank-content loss.} The generated image is put into the Ranker to predict the ranking score. Then, the rank-content loss can  be defined as:
\begin{equation}
\setlength{\abovedisplayskip}{4pt} 
\setlength{\belowdisplayskip}{4pt}
L_R=sigmoid(R(G(x_i))),
\end{equation}
where $ R(G(x_i)) $ is the ranking score of the generated image. A lower ranking score indicates better perceptual quality. After applying the sigmoid function, $ L_R $ represents ranking-content loss ranging from $0$ to $1$.
\begin{figure}[t]
\setlength{\abovecaptionskip}{-0.2cm}
\setlength{\belowcaptionskip}{-0.2cm}
\begin{center}
	\includegraphics[width=1\linewidth]{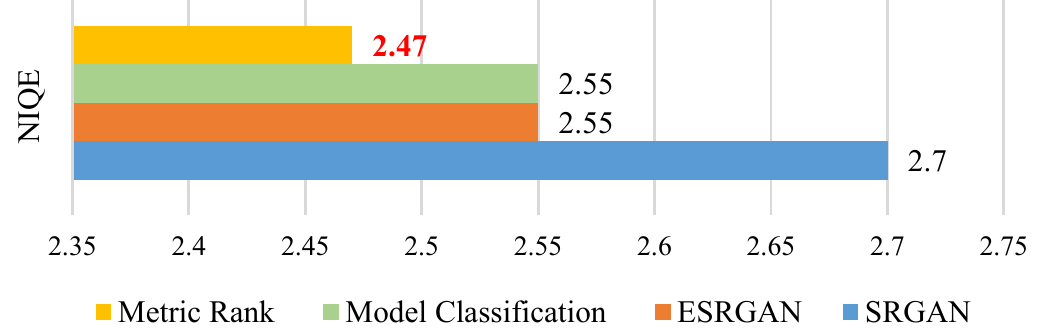} 

\end{center}
   \caption{The upper bound (average NIQE value) of SRGAN, ESRGAN, model rank and model classification.}
\label{fig:4}

\end{figure}

\subsection{Upper bound of RankSRGAN}
\label{3.4}
The proposed Ranker possesses an appealing property: by elaborately selecting the SR algorithms and the perceptual metric, the RankSRGAN has the potential to surpass the upper bound of these methods and achieve superior performance. To validate this comment, we select the state-of-the-art perceptual SR methods -- SRGAN \cite{ledig2017photo} and ESRGAN \cite{Wang_2018_ECCV_Workshops} to build the rank dataset.  Then we use the perceptual metric NIQE \cite{mittal2013making} for evaluation. NIQE is demonstrated to be highly correlated with human ratings and easy to implement. A lower NIQE value indicates better perceptual quality. When measured with NIQE on the PIRM-Test \cite{blau20182018} dataset, the average scores of SRGAN and ESRGAN are 2.70 and 2.55, respectively. ESRGAN obtains better NIQE scores for most images but not all images, indicating that SRGAN and ESRGAN have mixed ranking orders with NIQE.

In order to examine the effectiveness of our proposed Ranker, we compare two ranking strategies -- metric rank and model classification. Metric rank, which is our proposed method, uses perceptual metrics to rank the images. For example, in each image pair, the one with a lower score is labeled to 1 and the other is 2. The model classification, as the comparison method, ranks images according to the used SR methods, i.e., all results of ESRGAN are labeled to 1 and those of SRGAN are labeled to 2. We then give an analysis of the upper bound of these two methods. The upper bound can be calculated as:
\begin{small}
\begin{equation}
\setlength{\abovedisplayskip}{5pt}
\setlength{\belowdisplayskip}{5pt}
\begin{split}
UB_{MC}= Mean(PM_{SR2-L} + PM_{SR2-H} )\\
UB_{MR}= Mean(PM_{SR2-L} + PM_{SR1-L})\\
where: PM_{SR1-L} < PM_{SR2-H},
\end{split}
\label{equation:10}
\end{equation}
\end{small}where $UB_{MC}$ and $UB_{MR}$ represent the upper bound of model classification and metric rank, respectively. $PM$ (Perceptual Metric) is the perceptual score for each image in the corresponding class. (SR1, SR2) represents two SR results of the same LR. Subscripts $-L$ and $-H$ indicate the Lower and Higher perceptual score in (SR1, SR2). We use (SRGAN, ESRGAN) as (SR1, SR2) to obtain the upper bound of these methods, as shown in Figure \ref{fig:4}. Obviously, metric rank could combine the better parts of different algorithms and exceed the upper bound of a single algorithm.

We further conduct SR experiments to support the above analysis. We use the metric rank and model classification approach to label the rank dataset. Then the Ranker-MC (model classification) and Ranker-MR (metric rank) are used to train separate RankSRGAN models. Figure \ref{fig:5} shows the quantitative results (NIQE), where RankSRGAN-MR outperforms ESRGAN and RankSRGAN-MC. This demonstrates that our method could exceed the upper bound of all chosen SR algorithms.

\subsection{Comparison with Discriminator and Perceptual loss}
\label{3.5}
We analyse the differences among Ranker, discriminator and perceptual loss. Generally speaking, they are complementary loss functions for natural image generation. The perceptual loss focuses on feature-level similarity, thus prefers to generate complex textures, which may not be natural-looking. To force the generated images to lie in the natural image space, the discriminator is proposed as an additional constraint. {Following SRGAN \cite{ledig2017photo} and ESRGAN \cite{Wang_2018_ECCV_Workshops}, we make a qualitative illustration in qualitative and theoretical analysis, since the optimization and implementation in practice of learning-based methods are quite empirical and technical.}
As shown in Figure \ref{fig:rankdata}, the intersection of VGG loss and Discriminator stopping space is the model output, i.e., $x_{SR_1}, x_{SR_2}$. Using different levels of feature maps (VGG22, VGG54) will also generate different results. By combining perceptual loss and discriminator, we could obtain visually pleasing images. However, these images may not have better perceptual index measured by objective metrics. ESRGAN introduces the relativistic discriminator to improve the perceptual index, but the promotion is not guaranteed to lie in the direction of perceptual metrics (see $x_{SR_3}$ and the curved line in Figure \ref{fig:rankdata}). The proposed Ranker acts as a complementary constraint of discriminator and perceptual loss. It optimizes the results in the direction of perceptual metrics by using hierarchical data $x_1, x_2,..., x_i$. The final result $x_{SR_4}$ will achieve better perceptual scores than $x_{SR_1}, x_{SR_2}, x_{SR_3}$. 

{SRGAN uses PSNR and MOS to evaluate the impact of VGG loss with different feature extractors. The shallower feature extractor tends to achieve better PSNR performance while the deeper one favors the perceptual quality. ESRGAN employs Relativistic Discriminator to enhance the perceptual quality of super-resolved results. However, its training procedure is not stable to a specific perceptual metric. This makes it hard to select the best model among different iterations. Therefore, we present the convergence curves to compare different designs for the specific perceptual metric.}

{
We use NIQE to compare different methods on perceptual metric. As we can see from Figure \ref{fig:1}, SRGAN-vgg22 that adopts perceptual loss with VGG22 layer tends to produce higher NIQE value because VGG22 layer extracts the low-level feature information. SRGAN-vgg54 presents obvious improvement in NIQE metric compared with SRGAN-vgg22. The VGG54 layer focuses on high-level information to generate perceptually pleasing results, but it is hard to continually improve NIQE performance by artificially adjusting VGG feature extractor. 
Although ESRGAN achieves better NIQE performance compared with SRGAN-vgg54 in average, there exits great fluctuation in model performance, which makes it difficult to select the best performance model on NIQE. As shown in Figure \ref{fig:onecol}, the training curve of ESRGAN presents significant oscillation in terms of NIQE among different iterations. The proposed RankSRGAN can achieve a more stable optimization process and obtain the best performance on perceptual metric.
}

\begin{figure}[t!]
\setlength{\abovecaptionskip}{-0.3cm}
\setlength{\belowcaptionskip}{-0.2cm}
\begin{center}
	\includegraphics[width=1\linewidth]{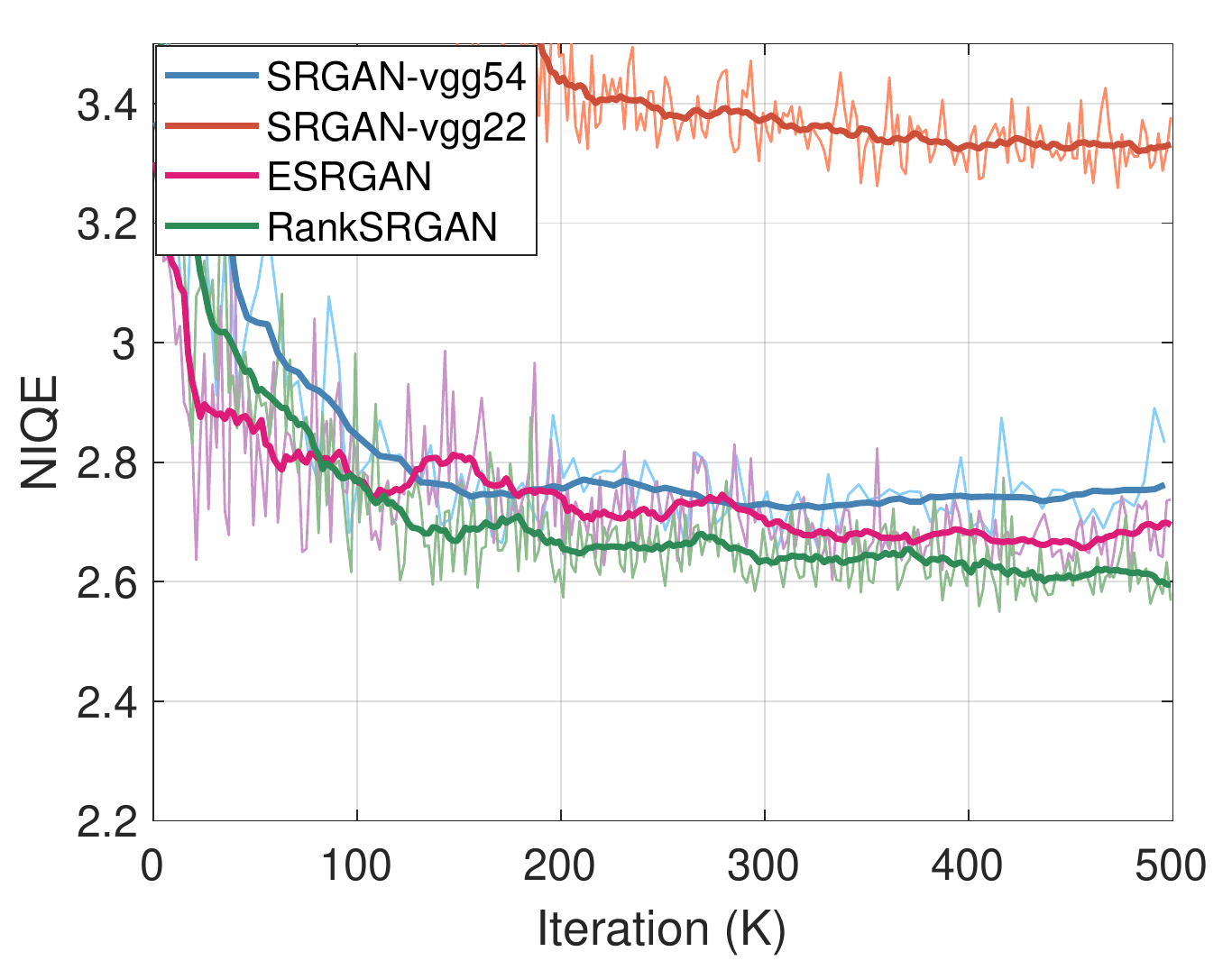} 

\end{center}
   \caption{The comparison of RankSRGAN and the state-of-the-art perceptual SR methods. NIQE: lower is better.}
\label{fig:1}
\label{fig:onecol}
\end{figure}

\begin{figure*}
\begin{center}
\includegraphics[width=1\linewidth]{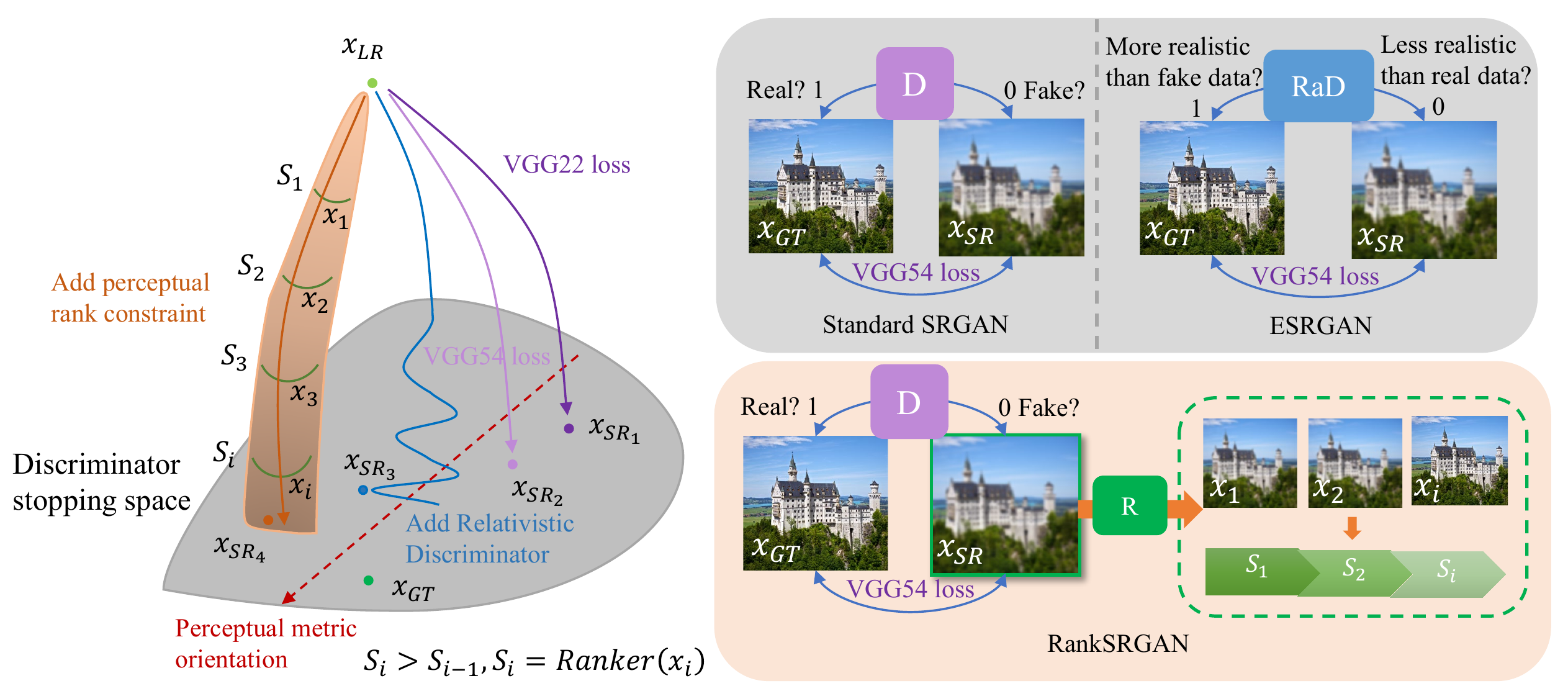} 
\end{center}
   \caption{Image space visualization. A generative network can yield different texture results by different constraints. In the standard SRGAN, VGG54 loss with high-level feature tends to generate better textual details while VGG22 loss with low-level features will produce to more smooth output. ESRGAN employs the relativistic discriminator to expand the solution space and generates SR results with better perceptual scores. However, existing methods just provide a single objective target. In the proposed RankSRGAN, we introduce Ranker to provide stage-wise guidance for generator. The Ranker could predict the scores of middle stages during the training procedure. Furthermore, the ranking score is defined as a rank-content loss to constrain the generator in the orientation of perceptual metrics. }
   
\label{fig:rankdata}
\end{figure*}

\begin{figure*}[ht!]
\centering
\subfigure[Rank datasets with different distortions]{\includegraphics[width=0.46\linewidth]{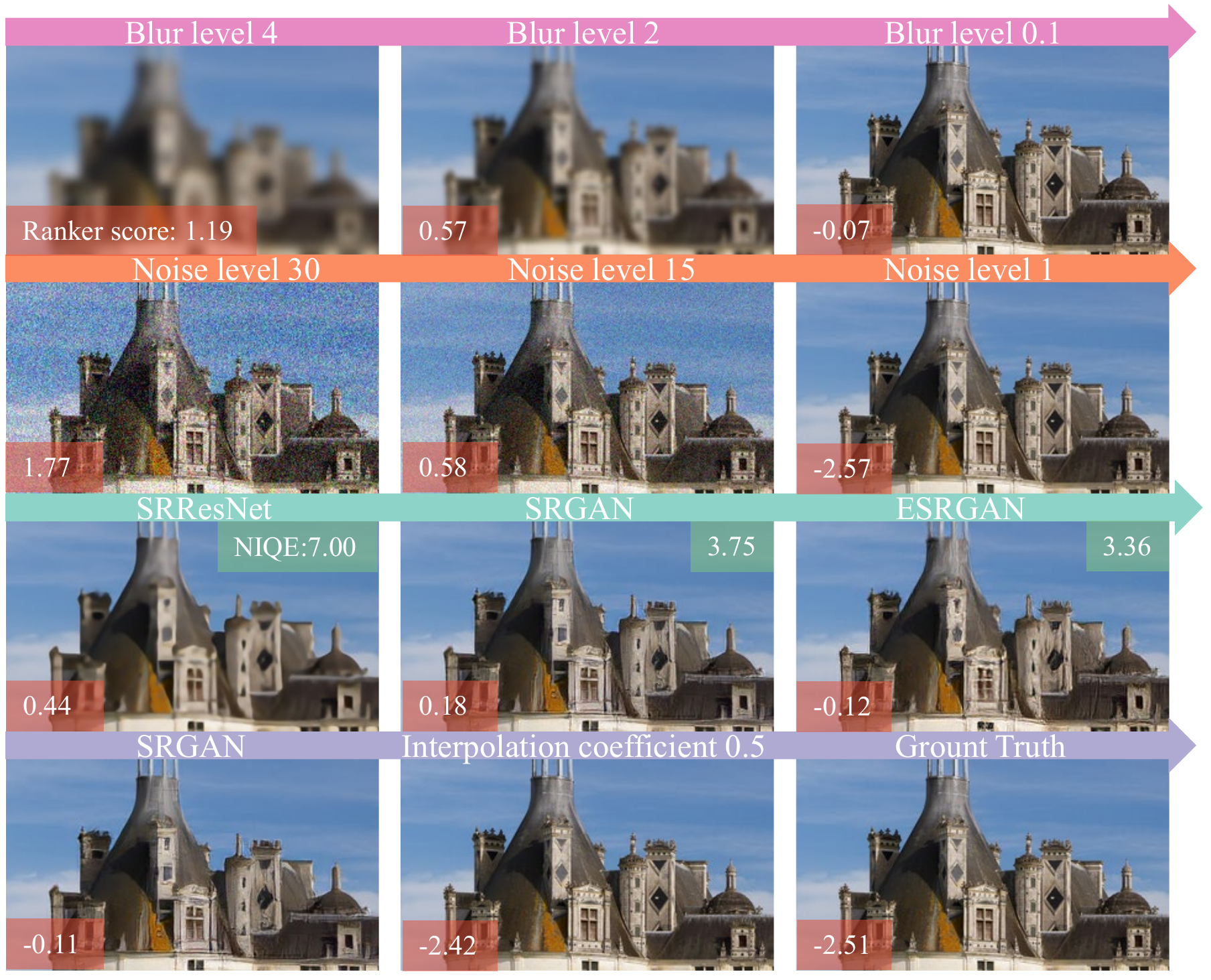}}
\quad
\subfigure[The convergence curves in NIQE]{\includegraphics[width=0.48\linewidth]{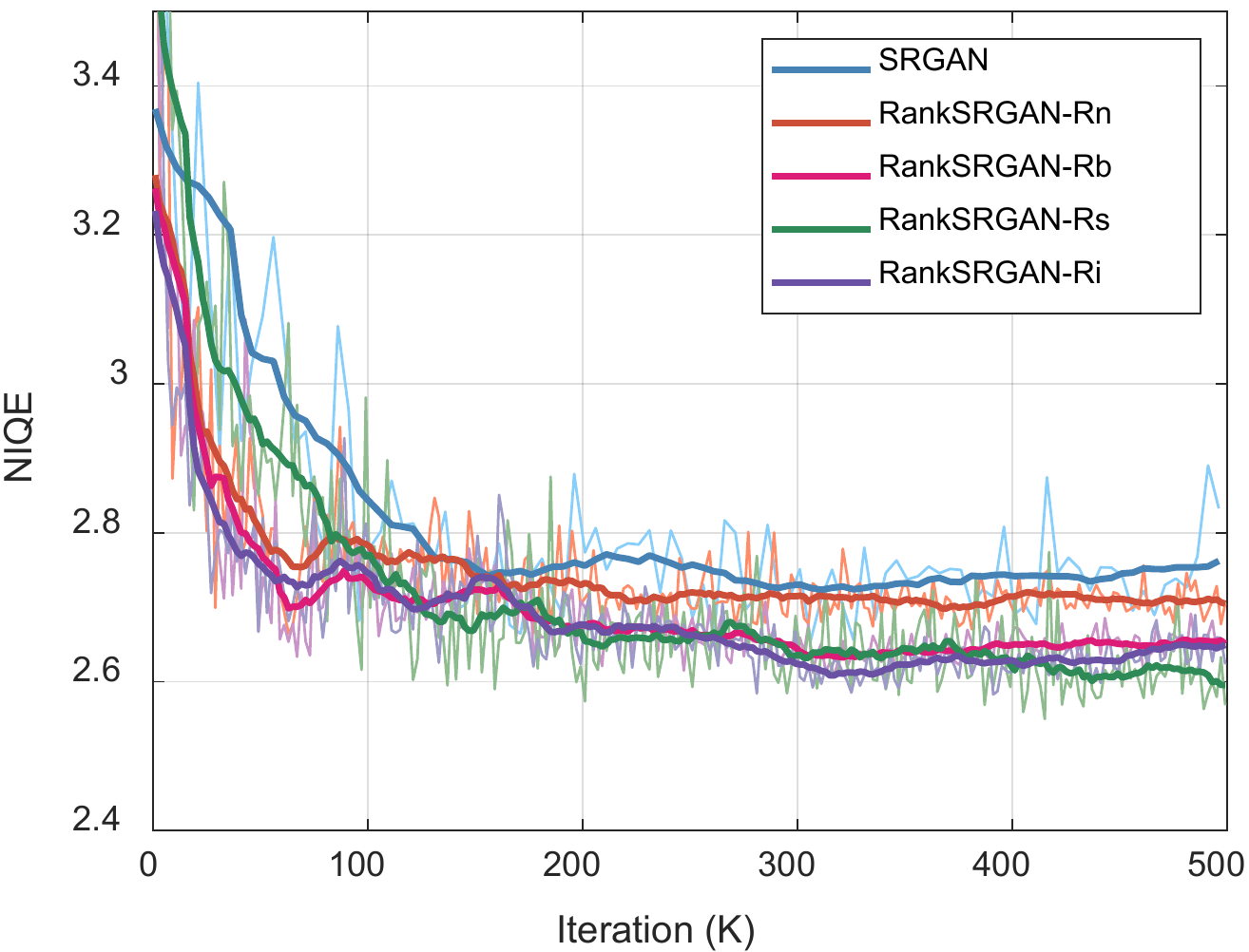}}

\caption{ The effects of Ranker with different distortions. (a) Different rank datasets with image distortion (blur and noise), SR and interpolation. For rank dataset with SR, we employ NIQE as the perceptual metric to get the order of pair-wise images. For rank datasets with blur, noise and image interpolation, we directly use degradation levels and interpolation coefficients to generate rank labels. The well-trained Ranker with different rank datasets can generate images with different perceptual characteristics. (b) The convergence curves of RankSRGAN with different Rankers. }
\label{fig:RankSRGANIQA}

\end{figure*}

Although higher perceptual index does not absolutely equal to better visual quality, the proposed Ranker provides a feasible way to tune the optimization direction of image generation. We could generate different kinds of images with different metrics to meet diverse requirements. The visual quality will also be improved along with the development of perceptual metrics.

\section{Extension to General Rank Datasets}
\label{section:rankdataset}

First, let us revisit why we use different SR results to generate the current rank dataset. Then we show that the rank dataset can be extended to a more general form. In this general design, we progressively exclude the use of SR algorithms and specific perceptual metrics, and propose two types of new rank datasets using image interpolation and distortion.    

\subsection{Rank Dataset with SR results}

The utilization of SR results is inspired by the recent progress of perceptual-oriented SR methods\cite{ledig2017photo,Wang2018ESRGAN}, which seems to be good guidance for further optimization. To construct a hierarchical dataset, we select three SR algorithms — SRResNet\cite{ledig2017photo}, SRGAN\cite{ledig2017photo} and ESRGAN\cite{Wang2018ESRGAN}. Next, we will explain why we select these three methods. To start with, the basic model adopted in this paper is SRGAN. Adding SRGAN results in the rank dataset could tell the Ranker ``where is the starting point''. We name this data as $\alpha_{base}$. Then we use a better algorithm — ESRGAN as positive guidance for Ranker. ESRGAN results can be recorded as $\alpha_{pos}$. We further require a negative example $\alpha_{neg}$ to prevent the output from approaching the opposite direction. SRResNet, as the generator of SRGAN, is optimized with MSE loss. If the Ranker neutralizes the effects of GAN loss and perceptual loss, the model will output SRResNet-like images. Thus SRResNet is a natural negative example or baseline for SRGAN. With these three algorithms, the Ranker could learn ``where to start and where to go''. More formally, we could define a standard rank dataset using the following formation: 
\begin{equation}
\begin{aligned}
\mathcal D = \{\alpha_i^{pos}, \alpha_i ^{base}, \alpha_i ^{neg}; PM(\alpha_i^{neg}), PM(\alpha_i ^{base}), PM(\alpha_i ^{pos})\}_{i=1}^N.
\end{aligned}
\label{equ:rankdata}
\end{equation}
Here $\mathcal{D}$ is the rank dataset, $N$ is the number of reference images, $PM$ is the score of the selected perceptual metric. With the ranking information, we can construct a pair-wise dataset with rank labels. {One merit of the Ranker is that it does not require tremendous data, but several representative and distinguishable classes. Because the learning procedure of the Ranker focuses on the orders of different categories rather than the actual value, which is quite different from the conventional regressor. From this point of view, Ranker can be regarded as a type of classifier. Hence, the number of categories does not determine its classification effect. We just need adequate data for each category (positive and negative examples) and this is much easier to obtain according to our approach.}

\textbf{Remark.} {Note that, in the formulation of constructing rank dataset, the selected perceptual metric is not constrained to a single one but can be any combination of multiple metrics. For example, perceptual index (PI) \cite{blau20182018} is a combination of NIQE \cite{mittal2013making} and Ma \cite{ma2017learning}, which is formulated as $PI = \frac{1}{2}((10-Ma)+NIQE)$. In Sec. \ref{section4.5}, experiments on PI and other metric combinations are conducted, showing that the proposed framework is capable of handling with combinations of different metrics.}

\subsection{ Rank Dataset with Image Interpolation}
According to the above analysis, we need a positive example and a negative example to constrain the proposed Ranker. The negative example is easy to obtain, but the positive example does not always exist. What if there are no better SR algorithms? The most straightforward way is to use the ground truth HR images directly. We have also conducted experiments in Section 5.4 to show the performance and differences in using HR and ESRGAN. Although the ground truth seems to be a perfect positive example, it may not be suitable for Ranker. The reason is that the proposed Ranker requires stage-wise guidance (see Figure 5), not only the final target. To bridge the gap between HR images and the SRGAN results, we need some intermediate data to play the role of the middle stages. Inspired by \cite{Wang_2018_ECCV_Workshops}, we apply image interpolation to generate intermediate results. The working mechanism is depicted in the following equation.
\begin{equation}
\begin{aligned}
I_{mid} = \lambda \dot I_{left}+(1-\lambda) \dot I_{right}, 0\leq \lambda \leq 1,
\end{aligned}
\label{equ:rankdata}
\end{equation}
where $I_{left}, I_{right}$ represent SRGAN and HR images, respectively. $\lambda$ is the coefficient. Image interpolation is a pixel-wise weighted sum of two images, which have the same content but different imagery effects. By setting $\lambda=0.25, 0.5, 0.75$, we can get a serious of intermediate data. As shown in Figure \ref{fig:RankSRGANIQA} (a), these images are changing continuously from a SRGAN output to an HR image. In addition, the training sample points can easily be further expanded to more data points in the proposed rank dataset. To decrease training cost, we do not adopt more training sample points due to three data sample points have achieved more better results that SRGAN. Experiments in Section \ref{Multiple Rankers} shows that the model trained with image interpolation strategy could also achieve better NIQE/PSNR values than SRGAN.

\subsection{ Rank Dataset with Distortion}
\begin{figure*}[htbp]
\setlength{\abovecaptionskip}{-0.2cm}
\setlength{\belowcaptionskip}{-0.2cm}
\begin{center}
	\includegraphics[width=1\linewidth]{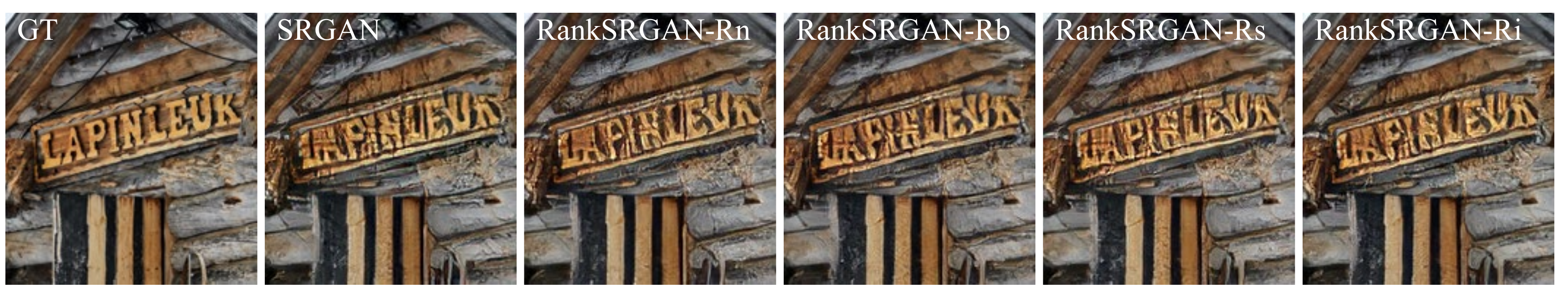} 
\end{center}
   \caption{The visual results of RankSRGAN with different rank dataset. [Rn, Rb, Rs, Ri]: Ranker trained by rank dataset with [noise distortion, blur distortion, image SR and image interpolation].}
\label{fig: evaluation2}

\end{figure*}

The above two rank datasets still require SRGAN results. In essence, as long as we can get hierarchical images with rank labels, it is unnecessary to use SR methods or even perceptual metrics. One way to generate this kind of dataset is to introduce artificial distortions on the ground truth HR images. This is a customized operation in the generation of IQA benchmark. For instance, TID2013 dataset \cite{2015Image} applies 24 different distortions (e.g., Gaussian blur, Gaussian noise, mean shift, and etc.) on 25 reference images. Similarly, we can add different levels of distortions on an image to generate hierarchical data. In this work, we use 3650 images in DF2K dataset (DIV2K+Flickr2K) as reference images and Gaussian blur/noise as the distortion types. To generate different levels, we use 3 blur kernels (width 0.1, 2, 4) for Gaussian blur and 3 covariances (1, 15, 30) for Gaussian noise. Obviously, there is no need to use an objective metric to evaluate its perceptual quality. We can directly rank these images according to their distortion levels. In Figure \ref{fig:RankSRGANIQA} (a), we show some examples of using Gaussian blur/noise. We can observe that the learned ranking scores are in accordance with NIQE index and distortion levels. In Figure \ref{fig:RankSRGANIQA} (b), we compare the RankSRGAN with different rank datasets. Generally, their performance can be ranked as ``SR methods $>$ image interpolation $>$ distortion''. Furthermore, different rank datasets will also generate images with different perceptual characteristics. For example, using Gaussian blur will produce sharp textures, while using image interpolation will favour images with less GAN artifacts. These visual differences are shown in Figure \ref{fig: evaluation2}. Detailed experimental settings and comparisons are presented in Section \ref{section: General Rank}. 

\section{Experiments}
\subsection{Training details of Ranker}
\label{4.1}
\textbf{Datasets.} We use DIV2K training set (800 images) \cite{agustsson2017ntire} and Flickr2K (2650 images) \cite{timofte2017ntire} dataset to generate pair-wise images as rank dataset for training and validation. Specifically, we randomly split the whole 3450 images into two parts: $90\%$ for training and the remaining $10\%$ for validation. We evaluate the Ranker on this validation set.

Three different SR algorithms (SRResNet\cite{ledig2017photo}, SRGAN \cite{ledig2017photo} and ESRGAN \cite{Wang_2018_ECCV_Workshops}) are used to generate super-resolved images as three perceptual levels, as shown in Table \ref{table:1}.

\begin{table}[!ht]
\caption{The performance of super-resolved results with three perceptual levels in PIRM-Test \cite{blau20182018}.}

\small 
\setlength{\abovecaptionskip}{-3pt}
\setlength{\belowcaptionskip}{-10pt}
\begin{center}
\begin{tabular}{c|ccc}
\hline\hline
PIRM-Test & SRResNet & SRGAN & ESRGAN \\
\hline
NIQE & 5.968 & 2.705 & 2.557\\
PSNR & 28.33 & 25.62 & 25.30\\
\hline
\end{tabular}
\end{center}

\label{table:1}
\vskip -0.4cm
\end{table}
We extract patches from those pair-wise images with a stride of $ 200 $ and size of $ 296\times 296 $. For one perceptual level (SR algorithm), we can generate 150 K patches (10$\%$ for validation, 90$\%$ for training). {Inspired by PIRM2018-SR Challenge\cite{blau20182018}, we adopt NIQE\cite{mittal2013making} as the perceptual metric. We use the official released codes to calcuate NIQE \footnote{https://github.com/roimehrez/PIRM2018}.} Other metrics will be investigated in Section \ref{section4.5}. Finally, we label every image pair to (3,2,1) according to the order of corresponding NIQE value (the one with the best NIQE value is set to 1).

\textbf{Implementation details.} As shown in Figure \ref{fig:2}, we  utilize VGG\cite{simonyan2014very} structure to implement the Ranker \cite{liu2017rankiqa}, which includes 10 convolutional layers, a series of batch normalization and LeakyReLU operations. Instead of max-pooling, we apply convolutional layer with a kernel size $4$ and stride $2$ to downsample the features. In one iteration, two patches with different perceptual levels are randomly selected as the input of Ranker. 
For optimization, we use Adam \cite{kingma2014adam} optimizer with weight decay $1\times10^{-4}$. The learning rate is initialized to $1\times10^{-3}$ and decreases with a factor $0.5$ of every $10\times10^{4}$ iterations for total $30\times10^{4}$ iterations. The margin $\epsilon$ of margin-ranking loss is set to $0.5$. {Note that the margin-ranking loss is only used to train the Ranker.} For weight initialization, we use He. \cite{he2015delving} method to initialize the weights of Ranker.

\textbf{Evaluation.} The Spearman Rank Order Correlation Coefficient (SROCC) \cite{liu2017rankiqa} is a traditional evaluation metric to evaluate the performance of image quality assessment algorithms. In our experiment, SROCC is employed to measure the monotonic relationship between the label and the ranking score. Given $N$ images, the SROCC is computed as:
\begin{small}
\begin{equation}
SROCC=1 - \frac{6\sum_{i=1}^{N}(y_i-\hat{y_i})^2}{N(N^2-1)},
\end{equation}
\end{small}where $y_i$ represents the order of label, and $\hat{y_i}$ is the order of output score of the Ranker. SROCC has the ability to measure the accuracy of Ranker. The larger value of SROCC represents the better accuracy of Ranker. 

The architecture of Ranker is based on the VGG network \cite{simonyan2014very}. We train three VGG networks varying from shallow to
deep ones: VGG-8, VGG-12 and VGG-16. The details of architecture is presented in appendix. The {Table} \ref{table:Ranker} shows the number of parameters, and the performance in different models. For validation dataset, the Ranker-12 and Ranker-16 achieve a SROCC of ${0.88}$, which is an adequate performance compared with those in the related work \cite{choi2018deep, liu2017rankiqa}. Since the VGG-12 can achieve the same accuracy as VGG-16, we apply the VGG-8 and VGG-12 on RankSRGAN. 
\vskip -0.2cm

\begin{table}[h!]
\caption{The performance of Ranker with different network architectures.}
\small 
\setlength{\abovecaptionskip}{-3pt}
\setlength{\belowcaptionskip}{-10pt}
\begin{center}
\begin{tabular}{c|ccc}
\hline\hline
Model & VGG-8 & VGG-12 & VGG-16 \\
\hline
Number of params (K) & 7,069 & 13,734 & 19,194\\
SROCC & 0.83 & 0.88 & 0.88\\
\hline
\end{tabular}
\end{center}
\label{table:Ranker}
\vskip -0.7cm
\end{table}

%

\subsection{Training details of RankSRGAN}
\label{4.2}
We use the DIV2K \cite{agustsson2017ntire} official training dataset (800 images) to train RankSRGAN. The patch sizes of HR and LR are set to 296 and 74, respectively. For testing, we use benchmark datasets Set14 \cite{zeyde2010single}, BSD100 \cite{martin2001database} and PIRM-Test \cite{blau20182018}. PIRM-test is used to measure the
perceptual quality of SR methods in PIRM2018-SR \cite{blau20182018}. Following the settings of SRGAN\cite{ledig2017photo}, we employ a standard SRGAN \cite{ledig2017photo} as our base model. The generator is built with 16 residual blocks, and the batch-normalization layers are removed \cite{Wang_2018_ECCV_Workshops}. The discriminator utilizes the VGG network \cite{simonyan2014very} with ten convolutional layers. The mini-batch size is 8. At each training step, the combination of loss functions (Section \ref{3.3}) for the generator is:
\vskip -0.2cm
\begin{equation}
\setlength{\abovecaptionskip}{-0.4cm} 
\setlength{\belowcaptionskip}{-0.4cm}
L_{total}=L_P+0.005L_G+0.03L_R,
\end{equation}
where the weights of $L_G$ and $L_R$ are determined empirically to obtain high perceptual improvement\cite{choi2018deep,ledig2017photo,Wang_2018_ECCV_Workshops}. {Notably, the perceptual loss and adversarial loss are inherited from the SRGAN baseline to make fair comparison.}

The Adam\cite{kingma2014adam} optimization method with $ \beta_1=0.9 $ is used for training. For generator and discriminator, the initial learning rate is set to $1 \times 10^{-4} $ which is reduced by a half for multi-step $ [50\times10^3,100\times10^3,200\times10^3,300\times10^3]$. A total of $ 600\times10^3$ iterations are executed by PyTorch. 
In training procedure, we add Ranker to the standard SRGAN. The Ranker takes some time to predict the ranking score, thus the traning time is a little slower (about 1.18 times) than standard SRGAN\cite{ledig2017photo}. For the generator, the number of parameters remains the same as SRGAN\cite{ledig2017photo}.

\subsection{Comparison with the-state-of-the-arts}
\label{section4.3}
\begin{figure}[t]
	\setlength{\abovecaptionskip}{-0.3cm}
	\setlength{\belowcaptionskip}{-0.2cm}
	\begin{center}
		\includegraphics[width=0.9\linewidth]{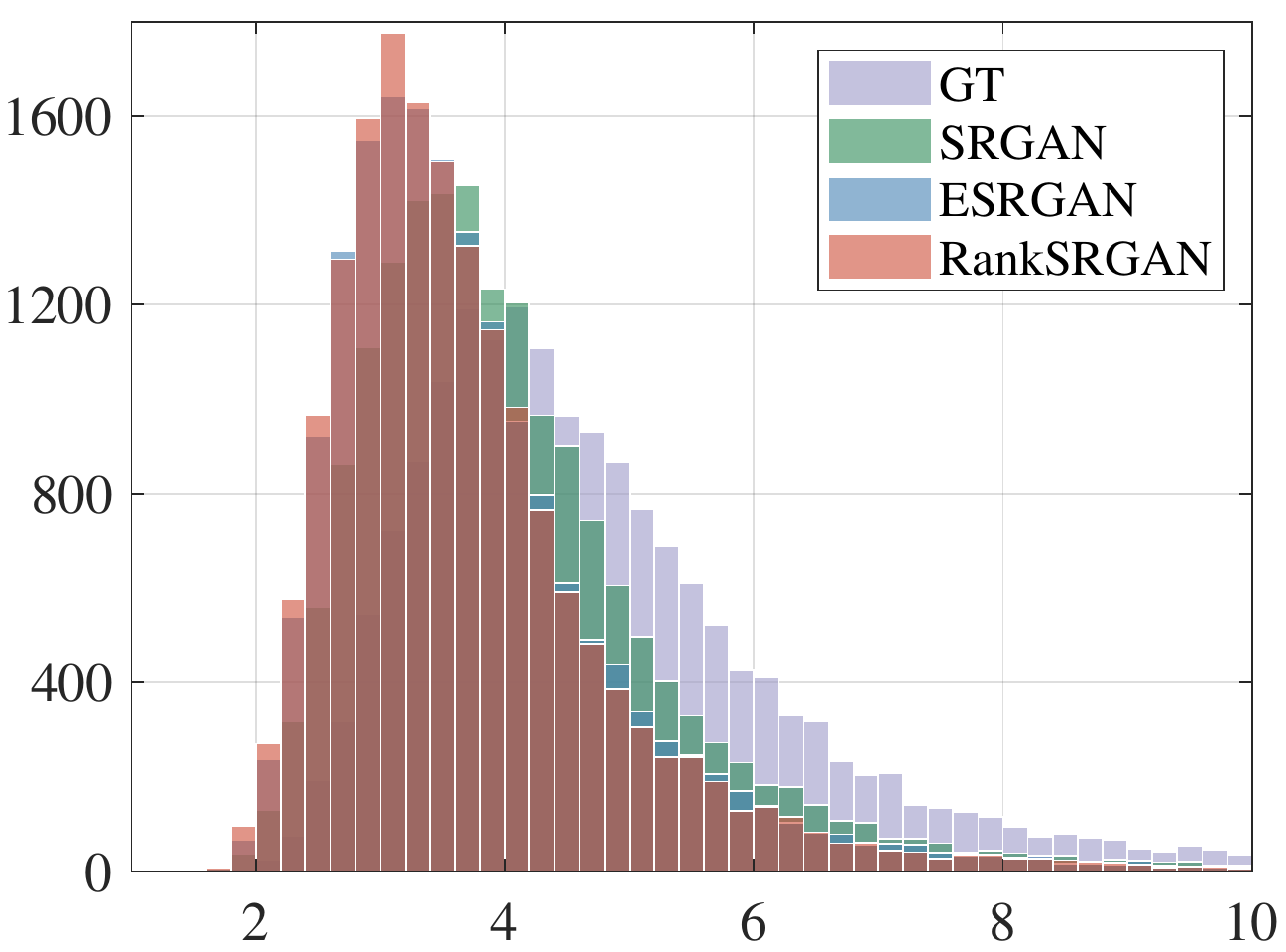} 
		
	\end{center}
	\caption{The histograms of NIQE score of GT, SRGAN, ESRGAN and RankSRGAN (ours). These graphs illustrate that RankSRGAN successfully achieves the best NIQE results. \textit{(Better viewed in color version)}}
	\label{fig:3}
	\label{fig:niqehist}
\end{figure}

We compare the performance of the proposed method with the state-of-the-art perceptual SR methods ESRGAN \cite{Wang_2018_ECCV_Workshops}/ SRGAN \cite{ledig2017photo} and the PSNR-orientated methods FSRCNN \cite{dong2016accelerating} and SRResNet \cite{ledig2017photo} \footnote{Our implementations of SRResNet and SRGAN achieve even better performance than that reported in the original paper.}. The evaluation metrics include NIQE \cite{mittal2013making}, PI \cite{blau20182018} and PSNR. Table \ref{table:2} shows their performance on three test datasets -- Set14, BSD100 and PIRM-Test. Note that lower NIQE/PI indicates better visual quality. When comparing our method with SRGAN and ESRGAN, we find that RankSRGAN achieves the best NIQE and PI performance on all test sets. Furthermore, the improvement of perceptual scores does not come at the price of PSNR. Note that in PIRM-Test, RankSRGAN also obtains the highest PSNR values among perceptual SR methods. Figure \ref{fig:Visual results} shows some visual examples, where we observe that our method could generate more realistic textures without introducing additional artifacts (please see the windows in Img 233 and feathers in Img 242). 

{In addition, we use DIV2K and Flickr2K validation set (as illustrated in Section \ref{4.1}) to evaluate the difference between GT images and GAN outputs. The NIQE score distribution is shown in Figure \ref{fig:niqehist}. We observe that most of GAN-based methods can achieve lower NIQE scores than GT images. Specifically, for SRGAN, ESRGAN and RankSRGAN, the proportion of their NIEQ scores lower than GT images is $76.2\%$, $86.6\%$ and $90.4\%$, respectively. The average NIQE score of GT, SRGAN, ESRGAN and RankSRGAN is 4.90, 4.12, 3.81 and 3.74, respectively. (Lower NIQE indicates better.) RankSRGAN achieves the best average NIQE score than the GT images.}

As the results may vary across different iterations, we further show the convergence curves of RankSRGAN in Figure \ref{fig:s5curve}. Their performance on NIQE and PSNR are relatively stable during the training process. For PSNR, they obtain comparable results. But for NIQE, RankSRGAN is consistently better than SRGAN by a large margin. 

\begin{figure}[htbp]
	\setlength{\abovecaptionskip}{-0.3cm}
	\setlength{\belowcaptionskip}{-0.2cm}
	\begin{center}
		\includegraphics[width=0.9\linewidth]{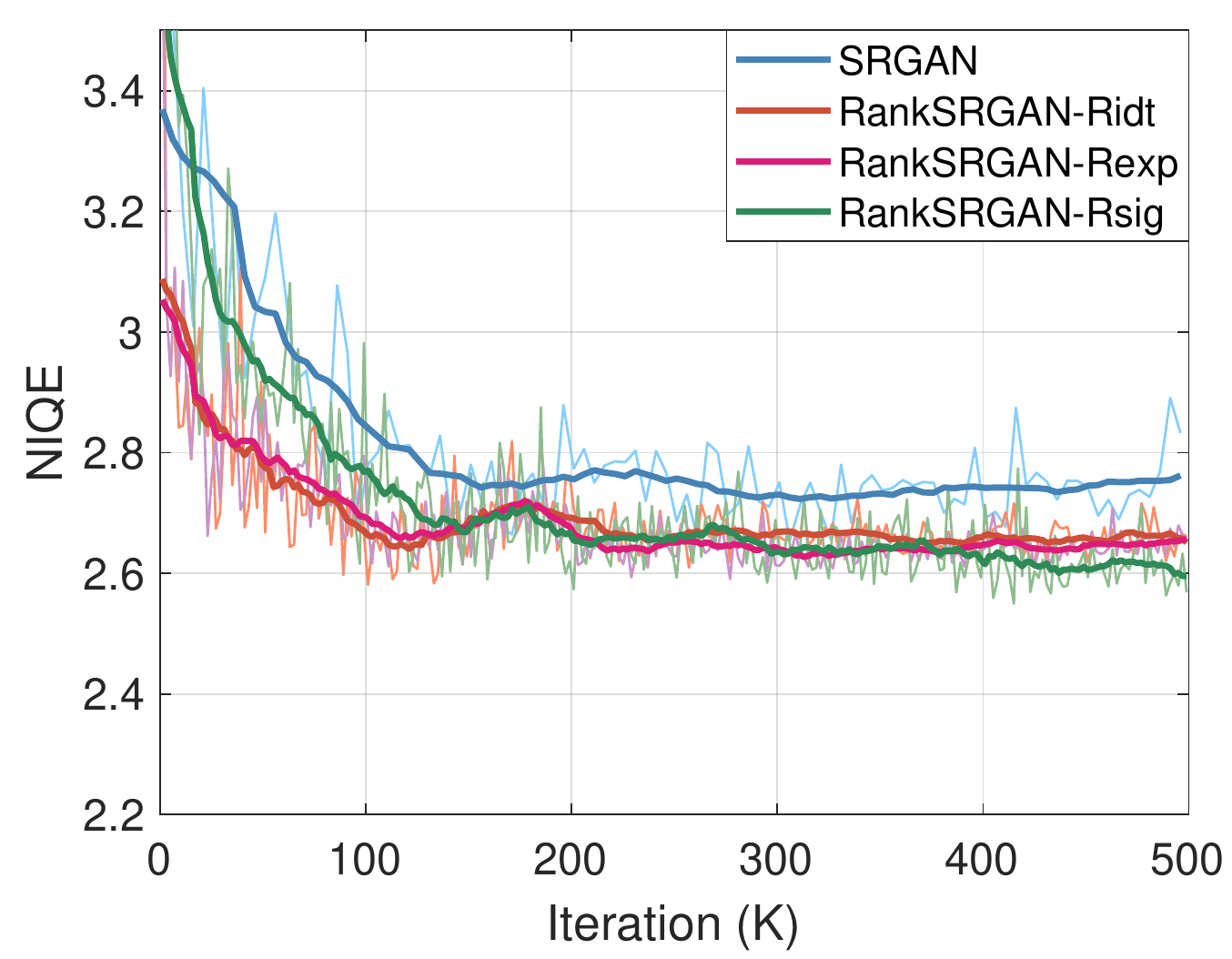} 
		
	\end{center}
	\caption{The comparison of RankSRGAN and the state-of-the-art perceptual SR methods on $\times$4. NIQE: lower is better.}
	\label{fig:function}
\end{figure}

\begin{table*}
\caption{\label{table:2}Average NIQE \cite{mittal2013making}, PI \cite{blau20182018} and PSNR values on the Set14 \cite{zeyde2010single}, BSD100 \cite{martin2001database} and PIRM-Test \cite{blau20182018}.}

\setlength{\abovecaptionskip}{0.2cm}
\setlength{\belowcaptionskip}{-0.8cm}
\begin{center}
\begin{tabular}{c|c|cccccc}
\hline\hline
Dataset & Metric & Bicubic & FSRCNN & SRResNet & SRGAN & ESRGAN & RankSRGAN (ours)\\
\hline
\multirow{3}*{Set14}&NIQE & 7.61 &6.92&6.12&3.82&3.28&\textbf{3.28}\\
       & PI & 6.97 & 6.16 &5.36&2.98&2.61&\textbf{2.61}\\
			   & PSNR & 26.08 & 27.66 &\textbf{28.57}&26.68&26.39&26.57\\
\hline
\multirow{3}*{BSD100}&NIQE & 7.60 &7.11&6.43&3.29&3.21&\textbf{3.01}\\
        & PI & 6.94 & 6.17 &5.34&2.37&2.27&\textbf{2.15}\\
	            & PSNR & 25.96 & 26.94 &\textbf{27.61}&25.67 &25.72 &25.57\\
\hline
\multirow{3}*{PIRM-Test}&NIQE & 7.45 &6.86& 5.98&2.71&2.56& \textbf{2.51}  \\
           & PI & 7.33 & 6.02 &5.18&2.09&1.98&\textbf{1.95}\\
          		   & PSNR & 26.45 & 27.57 &\textbf{28.33}&25.60&25.30&25.62\\
\hline
\end{tabular}

\end{center}

\end{table*}


\begin{figure*}[h]
\setlength{\abovecaptionskip}{-0.2cm}
\setlength{\belowcaptionskip}{-0.1cm}
\begin{center}
\includegraphics[width=1\linewidth]{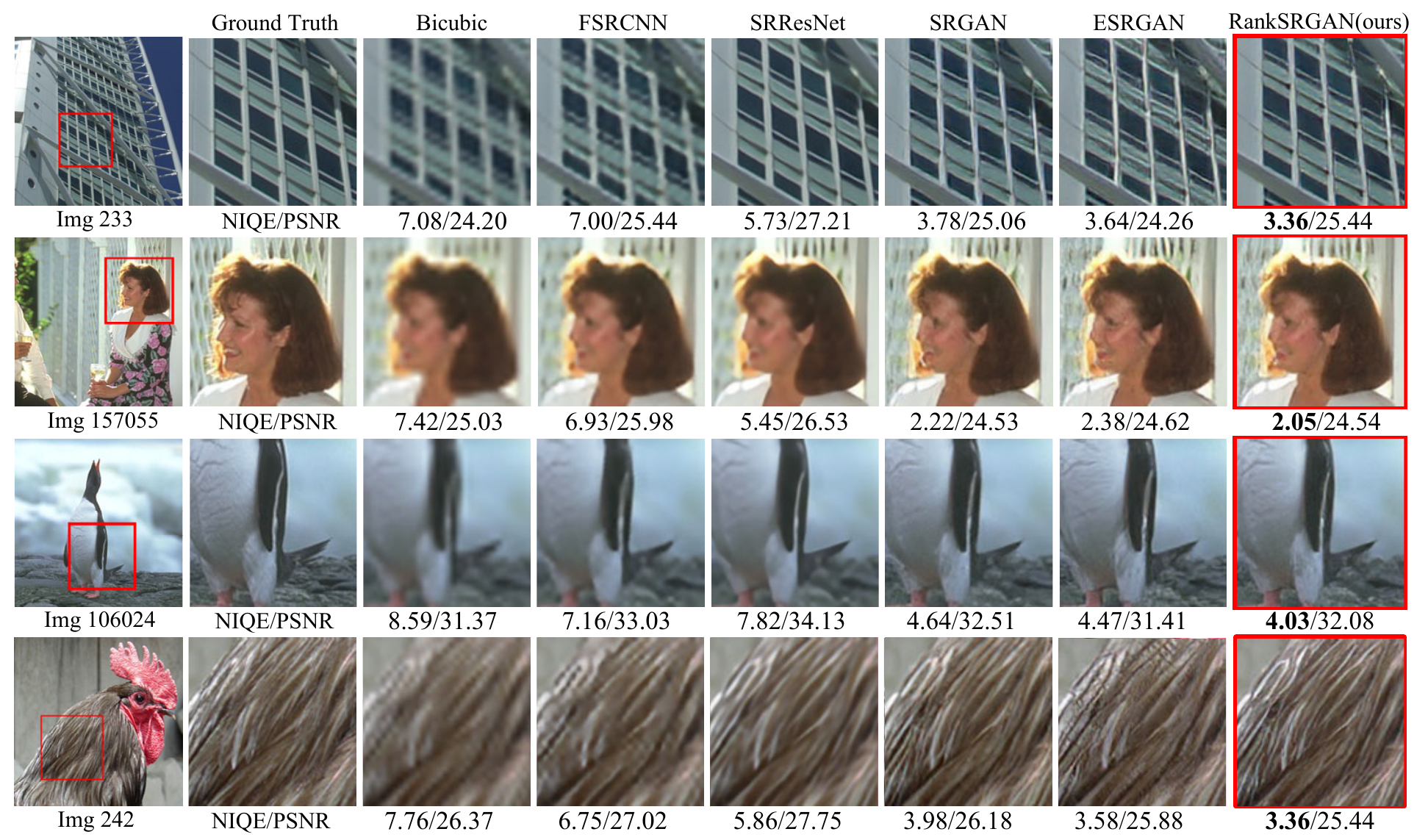} 
\end{center}
   \caption{Visual results comparison of our model with other works on $\times 4$ super-resolution. Lower NIQE value indicates better perceptual quality, while higher PSNR indicates less distortion.}
\label{fig:Visual results}
\vskip -0.5cm
\end{figure*}

\subsection{Ablation study}
\label{section4.5}

\textbf{Exploration of rank-content loss.} Further, we first test the monotonic function for obtaining rank-content loss from Ranker, since different monotonic functions may lead to different numerical outputs. We have tried a lot of monotonic functions, including Sigmoid, Exponential function and Identity function. The convergence curves are presented in Figure \ref{fig:function}. It is observed that the monotonic function only affects the converging process. All the different monotonic functions achieve similar results, which are all clearly better than the baseline SRGAN.

\textbf{Effect of different rank datasets.} The key factor that influences the performance of Ranker is the choice of SR algorithms. In the main experiments, we use (SRResNet, SRGAN, ESRGAN) to generate the rank dataset. Then what if we select other SR algorithms? Will we always obtain better results than SRGAN? To answer the question, we first analyze the reason of using these three algorithms, then conduct another experiment using a different combination. 

As our baseline model is SRGAN, we need the Ranker to have the ability to rank the outputs of SRGAN. Since the training of SRGAN starts from the pre-trained model SRResNet, the Ranker should recognize the results between SRResNet and SRGAN. That is the reason why we choose SRResNet and SRGAN. Then the next step is to find a better algorithm that could guide the model to achieve better results. We choose ESRGAN as it surpasses SRGAN by a large margin in the PIRM2018-SR challenge \cite{blau20182018}. Therefore, we believe that a better algorithm than SRGAN could always lead to better performance. 
\vskip -0.1cm

\begin{table}[h]

\renewcommand\tabcolsep{2.0pt}
\small
\setlength{\abovecaptionskip}{-0.2cm}
\setlength{\belowcaptionskip}{-0.3cm}
\caption{Comparison with RankSRGAN and RankSRGAN-HR.}
\begin{center}
\begin{tabular}{l|ccc}
\hline\hline
Method &SRGAN&RankSRGAN&RankSRGAN-HR \\
\hline
NIQE&2.70&\textbf{2.51}&2.58\\
PSNR&25.62&25.60&\textbf{26.00}\\
\hline
\end{tabular}
\end{center}

\label{table:RankSRGAN-HR}
\vskip -0.3cm

\end{table}

\begin{figure*}[ht!]
\setlength{\abovecaptionskip}{-0.2cm}
\setlength{\belowcaptionskip}{-1.8cm}

\begin{center}
	\includegraphics[width=0.9\linewidth]{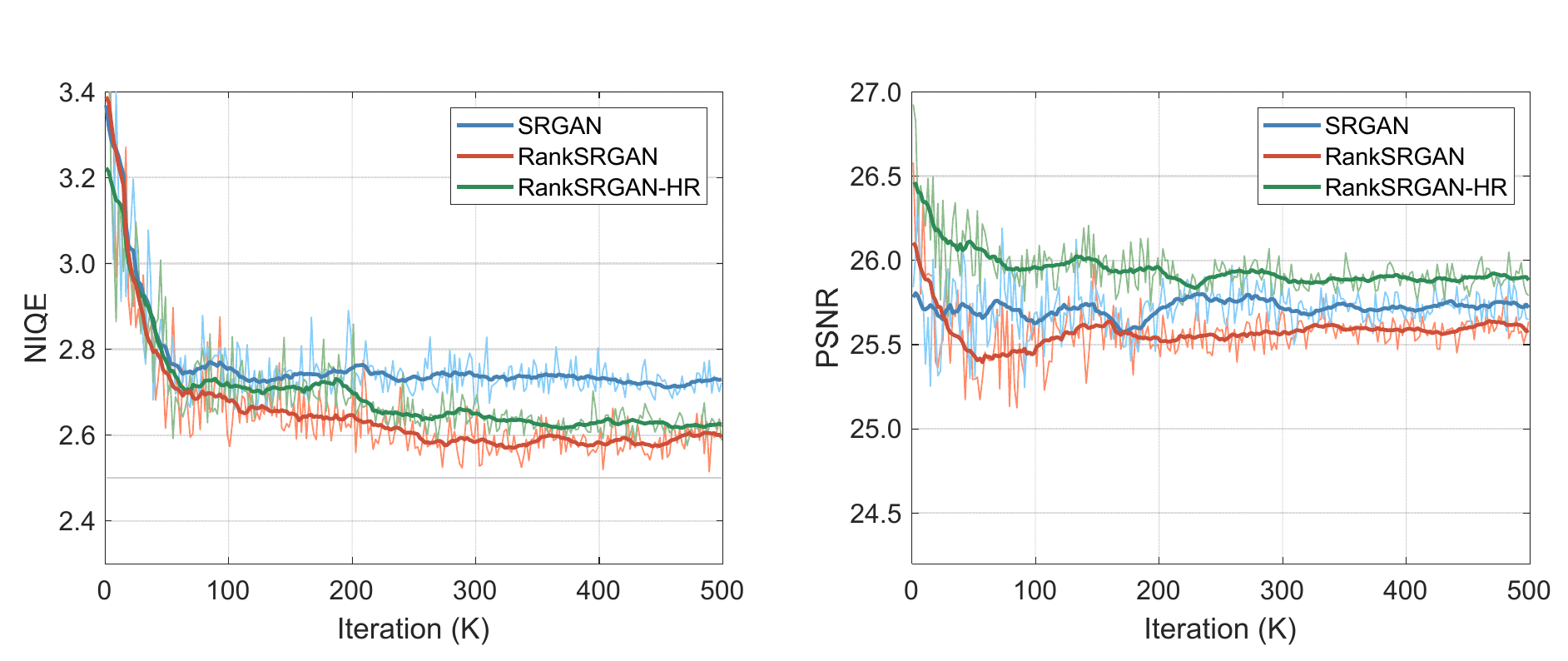} 
\end{center}
   \caption{Convergence curves of RankSRGAN and RankSRGAN-HR in PSNR/ NIQE.}
\label{fig:s5curve}

\end{figure*}

To validate this comment, we directly use the ground truth HR as the third algorithm, which is the extreme case. We still apply NIQE for evaluation. Interestingly, although HR images have infinite PSNR values, they cannot surpass all the results of SRGAN on NIQE. Similar to ESRGAN, HR and SRGAN have mixed ranking orders. We train our Ranker with (SRResNet, SRGAN, HR) and obtain the new SR model -- RankSRGAN-HR. Table \ref{table:RankSRGAN-HR} compares its results with SRGAN and RankSRGAN. As expected, RankSRGAN-HR achieves better NIQE values than SRGAN. But at the same time, RankSRGAN-HR also improves the PSNR by almost 0.4 dB. It achieves a good balance between the perceptual metric and PSNR. This also indicates that the model could always have an improvement space as long as we have better algorithms for guidance. 

\begin{figure*}[ht!]
\setlength{\abovecaptionskip}{-0.2cm}
\setlength{\belowcaptionskip}{-0.5cm}

\begin{center}
	\includegraphics[width=1\linewidth]{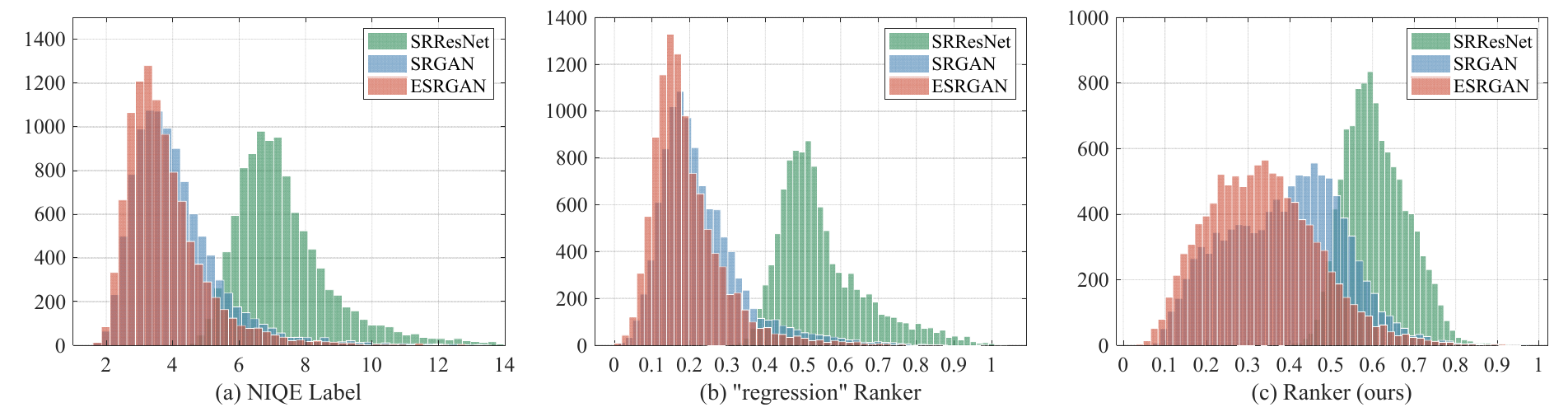} 
\end{center}
   \caption{The histograms of (a) NIQE label value, (b) the regression score of ``regression'' Ranker and (c) the ranking score of Ranker (ours). These graphs illustrate that Ranker (ours) successfully managed to separate the different perceptual levels. \textit{(Better view in color version)}}
\label{fig:histNIQE}
\end{figure*}

\begin{figure*}[ht!]
\setlength{\abovecaptionskip}{-0.2cm}
\setlength{\belowcaptionskip}{-0.5cm}

\begin{center}
	\includegraphics[width=1\linewidth]{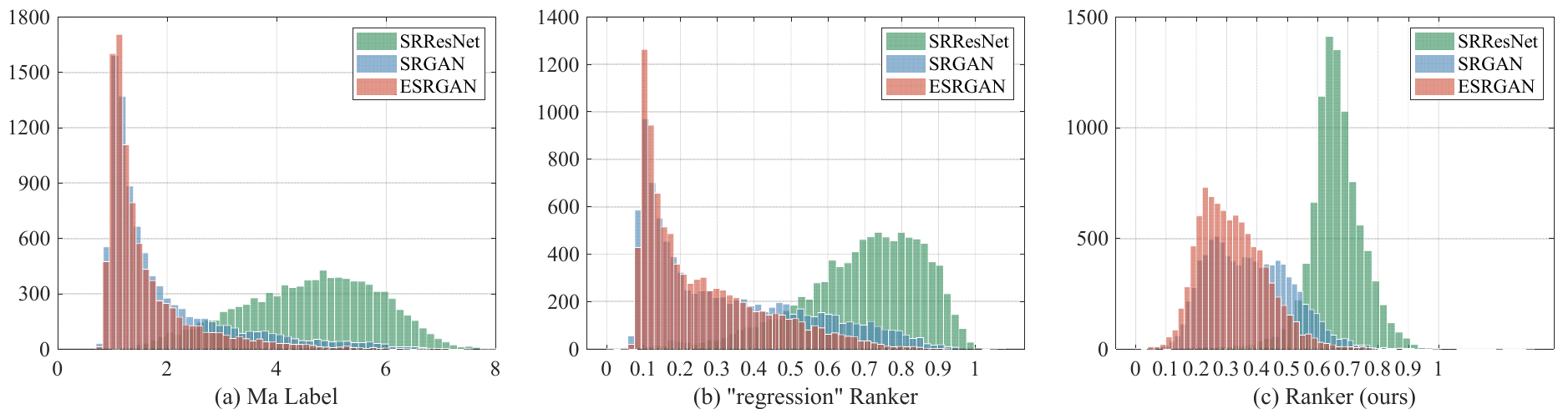} 
\end{center}
   \caption{The histograms of (a) Ma label value, (b) the regression score of ``regression'' Ranker and (c) the ranking score of Ranker (ours). These graphs illustrate that Ranker (ours) successfully managed to separate the different perceptual levels. \textit{(better view in color version)} }
\label{fig:histMa}
\end{figure*}

\textbf{Effect of Ranker: Rank VS. Regression.} To train our Ranker, we choose to use the ranking orders instead of the real values of the perceptual metric. Actually, we can also let the network directly learn the real values. In \cite{choi2018deep}, Choi et al. use a regression network to predict a subjective score for a given image and define a corresponding subjective score loss. To compare these two strategies, we train a ``regression'' Ranker with MSE loss instead of the margin-ranking loss. The labels in the rank dataset are real values of the perceptual metric. We use NIQE and Ma to generate the labels of rank dataset. All the other settings remain the same as RankSRGAN. 
\begin{table}[h]
\caption{The distance between $SR_1$ and $SR_2$ with regression and rank.}
\small 
\setlength{\abovecaptionskip}{-0.2cm} 
\setlength{\belowcaptionskip}{-0.4cm}
\begin{center}
\begin{tabular}{l|c c}
\hline\hline
Metric&Method&$E(|SR_1-SR_2|)$\\
\hline
NIQE&regression&0.06\\
NIQE&rank&\textbf{0.11}\\
\hline
Ma&regression&0.09\\
Ma&rank&\textbf{0.15}\\
\hline
\end{tabular}
\end{center}

\label{table:Ranker-distance}
\vskip -0.3 cm

\end{table}

To better understand the effects of Ranker with different learning strategy, we provide the histograms of NIQE /Ma label value in the validation dataset of the rank dataset. Furthermore, we plot the histograms of the output scores of different Rankers (“regression” Ranker and our Ranker) in Figure \ref{fig:histNIQE}. Comparing Figure \ref{fig:histNIQE} (b) and (c), Ranker successfully enlarges the distance between SRGAN and ESRGAN. The “regression” Ranker tends to learn the distribution of the NIQE label, while the NIQE values of SRGAN are close to ESRGAN. The same observation is also found in Ma metric as shown in Figure \ref{fig:histMa}.

\vskip -0.1cm

\begin{table}[h]
\caption{The performance of RankSRGAN with different Rankers. Re: Ranker with regression, $N$: Ranker with NIQE, $M$: Ranker with Ma.}
\small 
\setlength{\abovecaptionskip}{-0.1cm}
\setlength{\belowcaptionskip}{-0.3cm}
\begin{center}
\begin{tabular}{l|ccc}
\hline\hline
Method&NIQE&10-Ma&PSNR \\
\hline
SRGAN&2.71&1.47&25.62\\
ESRGAN&2.55&1.40&25.30\\
\hline
RankSRGAN-Re$_N$&2.53&1.42&25.58\\
RankSRGAN$_N$&\textbf{2.51}&1.39&25.60\\
\hline
RankSRGAN-Re$_M$&2.61&1.43&25.23\\
RankSRGAN$_M$&2.65&\textbf{1.38}&25.21\\
\hline
\end{tabular}
\end{center}

\label{table:RankSRGAN-regression}
\vskip -0.3cm

\end{table}

\begin{figure*}[t]
\setlength{\abovecaptionskip}{3pt}
\setlength{\belowcaptionskip}{3pt}
\begin{center}
	\includegraphics[width=1\linewidth]{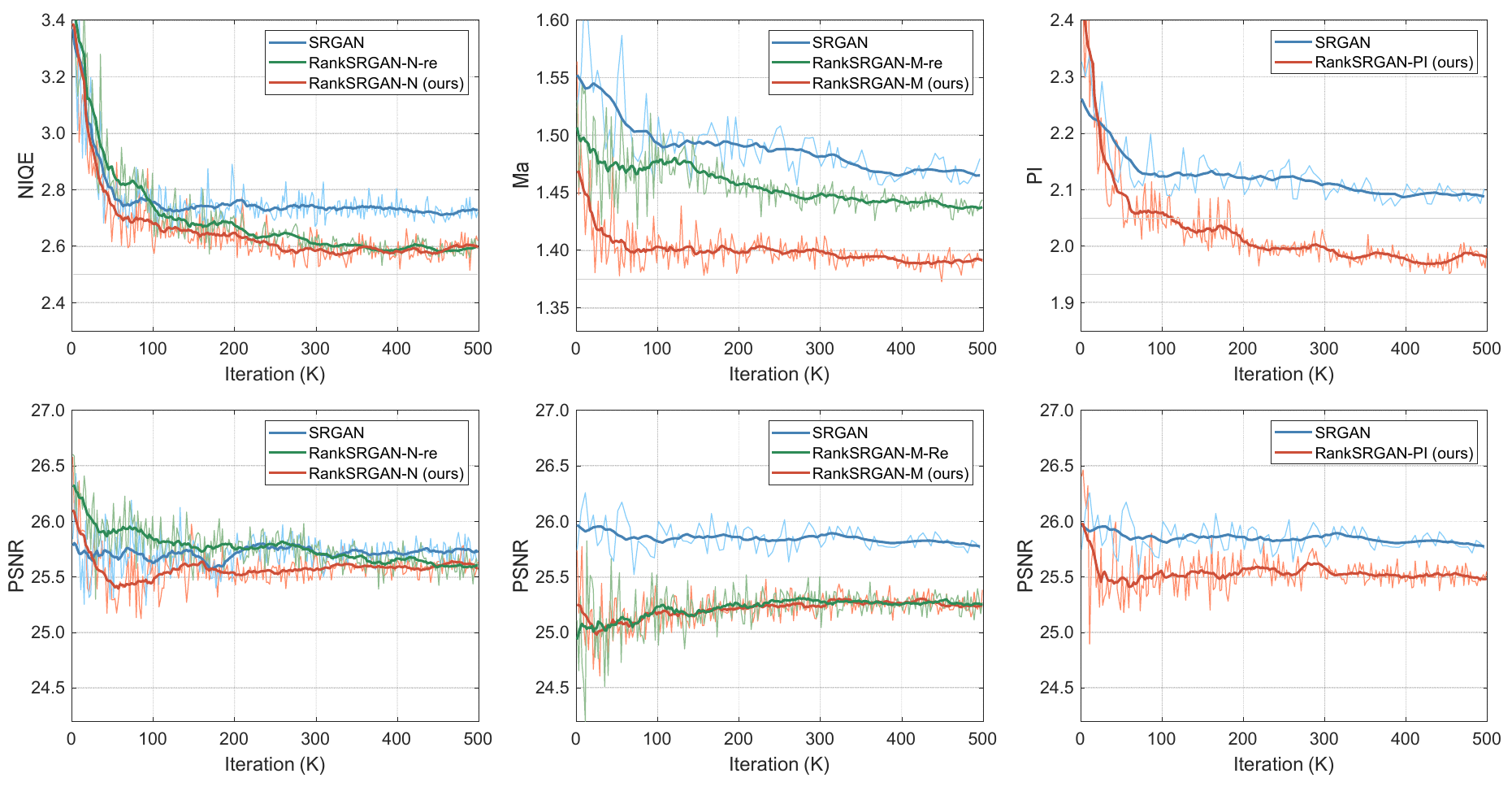} 

\end{center}
   \caption{The convergence curves of RankSRGAN$_N$, RankSRGAN$_M$ and RankSRGAN$_{PI}$.}
\label{fig:Metric curve}
\vskip -0.3 cm

\end{figure*}

Theoretically, the real values of perceptual metrics may distribute unevenly among different algorithms. For example, SRGAN and ESRGAN are very close to each other on NIQE values. This presents a difficulty for the learning of regression. On the contrary, learning ranking orders can simply ignore these variances. In experiments, we first measure the distances between the outputs of SRGAN and ESRGAN with different strategies. Table \ref{table:Ranker-distance} shows the mean absolute distances of these two strategies. Obviously, results with rank strategy have larger distances than results with regression. When applying these Rankers in SR training, the rank strategy achieves better performance than the regression strategy on the selected perceptual metric. Results are shown in Table \ref{table:RankSRGAN-regression}.

\textbf{Effect of different perceptual metrics.} As we claim that Ranker can guide the SR model to be optimized in the direction of perceptual metrics, we need to verify whether it works for other perceptual metrics. We choose Ma \cite{martin2001database} and PI \cite{blau20182018}, which show high correlation with Mean-Opinion-Score (Ma: 0.61, PI: 0.83) in \cite{blau20182018}. We use Ma and PI as the evaluation metric to generate the rank dataset. All other settings remain the same as RankSRGAN with NIQE. The only difference in these experiments is the ranking labels in the rank dataset. The results are summarized in Table \ref{table:RankSRGAN-metric}, where we observe that the Ranker could help RankSRGAN achieve the best performance in the chosen metric. This shows that our method can generalize well on different perceptual metrics.

\begin{figure*}[htbp]
\begin{center}
	\includegraphics[width=0.9\linewidth]{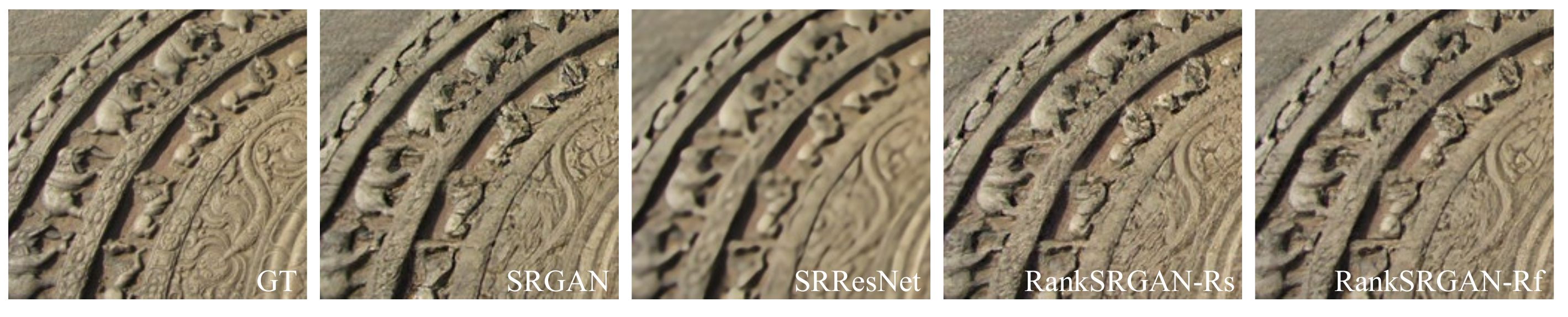} 

\end{center}
   \caption{The visual results of SRGAN and RankSRGAN with combination of different metrics. [Rs, Rf]: Ranker trained by rank dataset with NIQE and [NIQE, MSE].}
\label{fig:singleranker1}
\end{figure*}

\begin{figure*}[h!]
\begin{center}
	\includegraphics[width=0.9\linewidth]{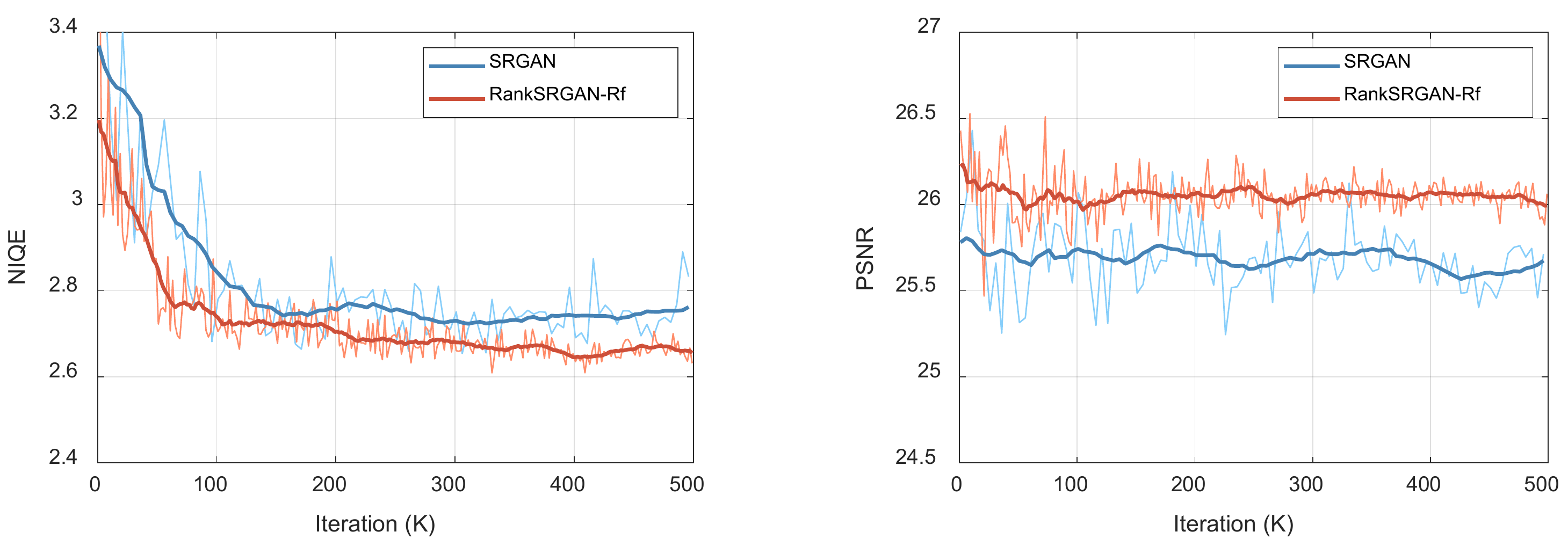} 

\end{center}
   \caption{The convergence curves of SRGAN and RankSRGAN-Rf. Rf: Ranker trained by fusion metric [NIQE, MSE]. }
\label{fig:singleranker2}
\end{figure*}

{Note that PI is actually a combination of two perceptual metrics (i.e., Ma and NIQE). In practice, any other combinations of measures can be treated in the same way. Here we further provide a toy example to validate such statements. Specifically, we employ the fused metrics ($NIQE+0.4*RMSE$) to generate the rank labels\footnote{Note that the sum of coefficients is not necessarily to be 1.}. The rank dataset uses SR algorithms as in Section 5.1. All the other settings remain the same as RankSRGAN. The quantitative results and qualitative examples are shown in Figure \ref{fig:singleranker1} and Figure \ref{fig:singleranker2} . We observe that the RankSRGAN with fused metrics could achieve improvement in both NIQE and PSNR orientation, compared with SRGAN. This demonstrates that the SR results can be controlled by multiple metrics. Notably, once we obtain the rank label, the Ranker is capable of learning the behavior of the combined evaluations and can further facilitate the subsequent optimization of SR network.}

\vskip -0.1cm

\begin{table}[h]
\begin{center}
\caption{The performance of RankSRGAN with different Rankers. $N$: Ranker with NIQE \cite{mittal2013making}, $M$: Ranker with Ma \cite{ma2017learning} and $PI$: Ranker with PI \cite{blau20182018}.}
\label{table:RankSRGAN-metric}

\small 
\setlength{\abovecaptionskip}{-0.3cm}
\setlength{\belowcaptionskip}{-0.4cm}
\begin{tabular}{l|cccc}
\hline\hline
Method &NIQE&10-Ma&PI&PSNR \\
\hline
SRGAN&2.71&1.47&2.09&25.62\\
ESRGAN&2.56&1.40&1.98&25.30\\
\hline
RankSRGAN$_N$&\textbf{2.51}&1.39&1.95&25.62\\
RankSRGAN$_M$&2.65&\textbf{1.38}&2.01&25.21\\
RankSRGAN$_{PI}$&2.49&1.39&\textbf{1.94}&25.49\\
\hline
\end{tabular}
\end{center}

\end{table}

We further present the curves showing that our RankSRGAN can achieve a constant improvement compared with the baseline SRGAN. We provide the curves of RankSRGAN-N, RankSRGAN-M, and RankSRGAN-PI (N: Ranker with NIQE,
M: Ranker with Ma, and PI: Ranker with PI) in Figure \ref{fig:Metric curve}. Besides, we add the curves of RankSRGAN-N-re and
RankSRGAN-M-re (re: “regression” Ranker). We observe that Ranker could help RankSRGAN achieve state-of-the-art
performance in the chosen metric. This shows that our method can generalize well on different perceptual metrics. Compared
with ”regression” Ranker, Ranker can accelerate the convergence of RankSRGAN-N. For RankSRGAN-M, Ranker can still
reach state-of-the-art performance (less than 1.40 in ESRGAN), while the “regression” Ranker cannot outperform ESRGAN
(1.40).

\begin{table}[h!]
\caption{The performance of RankSRGAN with the combination of loss functions (P: perceptual loss, R: rank-content loss, M: MSE loss). $\alpha_1, \alpha_2 : \left\{1,5\right\}$ }
\small 
\setlength{\abovecaptionskip}{-0.1cm} 
\setlength{\belowcaptionskip}{-0.4cm}
\begin{center}
\begin{tabular}{l|c|cc}
\hline\hline
Method & Loss &NIQE&PSNR \\
\hline
SRGAN&$L_P$&2.71&25.62\\
ESRGAN&$L_P+10L_M$&2.55&25.30\\

\hline
RankSRGAN&$L_P+L_{R}$&\textbf{2.51}&25.62\\
RankSRGAN-M$_1$&$L_P+L_{R}+\alpha_1L_M$&2.55&25.87\\
RankSRGAN-M$_2$&$L_P+L_{R}+\alpha_2L_M$&2.72&\textbf{26.62}\\

\hline
\end{tabular}
\vskip -0.3cm

\end{center}

\label{table:different loss}
\end{table}

\begin{figure*}[ht!]
\setlength{\abovecaptionskip}{-0.1cm}
\setlength{\belowcaptionskip}{-0.5cm}
\begin{center}

\subfigure[The results of user studies, comparing our method with SRGAN  \cite{ledig2017photo} and ESRGAN \cite{Wang_2018_ECCV_Workshops}.]{\includegraphics[width=0.45\linewidth]{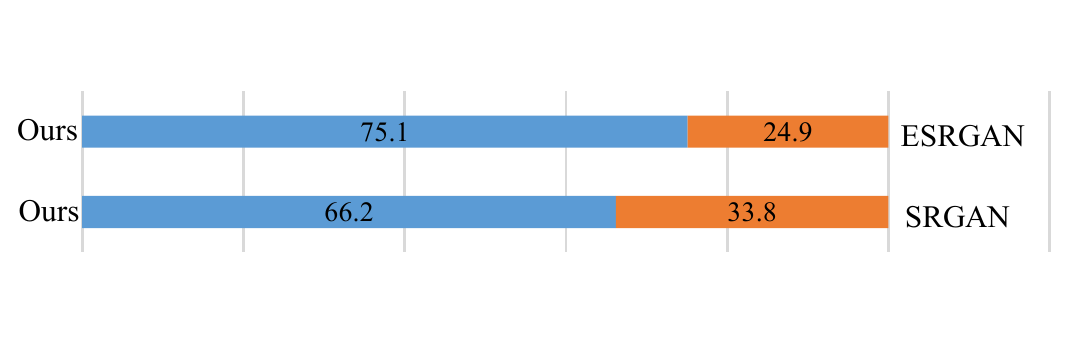}}
\quad
\subfigure[The ranking results of user studies: SRResNet \cite{ledig2017photo}, ESRGAN \cite{Wang_2018_ECCV_Workshops}, RankSRGAN (ours), and the original HR image.]{\includegraphics[width=0.45\linewidth]{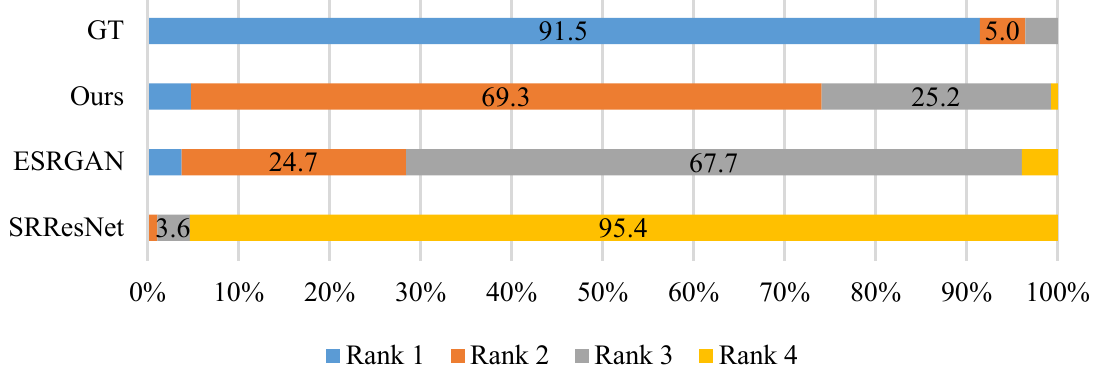}}

\end{center}
\caption{The results of user studies}
\label{fig:user study1}

\end{figure*}

Effect of different losses. To test the effects of rank-content loss, we expect to add MSE loss to achieve improvement in PSNR. Table \ref{table:different loss} shows the performance of our method trained with different combination of loss functions. 


As expected, increasing the contribution of the MSE loss with the larger $\alpha$ results in higher PSNR values. On the other hand, the NIQE values are increased which is a tradeoff between PSNR and NIQE as mentioned in \cite{blau2018perception}, and our method has a capability to deal with the priorities by adjusting the weights of the loss functions.

\subsection{User study}
\label{section: user study}

To demonstrate the effectiveness and superiority of RankSRGAN, we conduct a user study against state-of-the-art models, i.e. SRGAN \cite{ledig2017photo} and ESRGAN \cite{Wang_2018_ECCV_Workshops}. In the first session, two different SR images are shown at the same time where one is generated by the proposed RankSRGAN and the other is generated by SRGAN or ESRGAN. The participants are required to pick the image that is more visually pleasant (more natural and realistic). We use the PIRM-Test \cite{blau20182018} dataset as the testing dataset. There are a total of 100 images, from which 30 images are randomly selected for each participant. To make a better comparison, one small patch from the image is zoomed in. 

\begin{figure}[ht!]
\setlength{\abovecaptionskip}{-0.1cm}
\setlength{\belowcaptionskip}{-0.5cm}
\begin{center}
\includegraphics[width=0.9\linewidth]{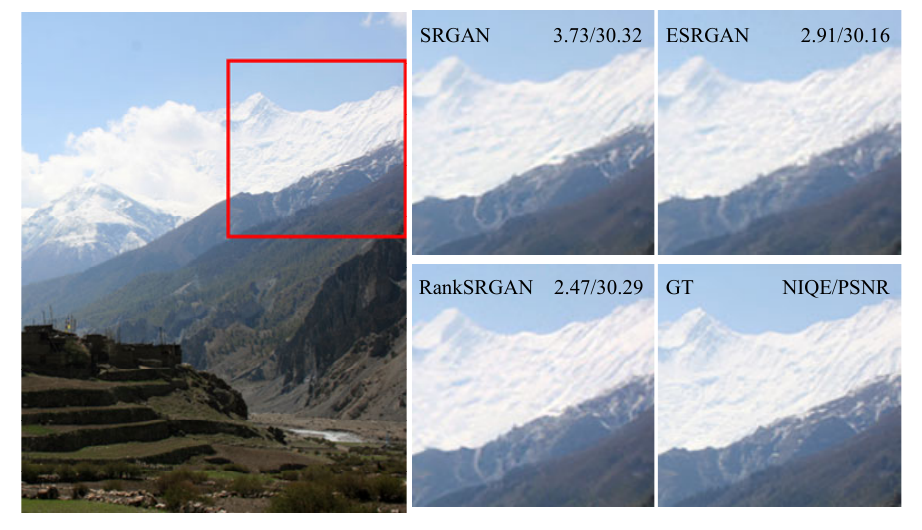} 
\end{center}
   \caption{The visual results of user studies, comparing our method with SRGAN  \cite{ledig2017photo} and ESRGAN \cite{Wang2018ESRGAN}.}
\label{fig: usvisual2}
\end{figure}

In the second session, we focus on the perceptual quality of different typical SR methods in a sorting manner. The participants are asked to rank 4 versions of each image: SRResNet \cite{ledig2017photo}, ESRGAN \cite{Wang_2018_ECCV_Workshops}, RankSRGAN, and the Ground Truth (GT) image according to their visual qualities. Similar to the first session, 20 images are randomly shown for each participant. There are totally 30 participants to finish the user study.

%

As suggested in Figure \ref{fig:user study1}, RankSRGAN has achieved better visual performance against ESRGAN and SRGAN. Since RankSRGAN consists of a base model SRGAN and the proposed Ranker, it can naturally inherit the characteristics of SRGAN and achieve better performance in perceptual metric. Thus, RankSRGAN performs more similar to SRGAN than ESRGAN. Figure \ref{fig: usvisual2} shows the ranking results of different SR methods. As RankSRGAN has the best performance in perceptual metric, the ranking results of RankSRGAN are second to GT images, but sometimes it even produces images comparable to GT. 

\begin{figure}[t!]
\setlength{\abovecaptionskip}{-0.2cm}
\setlength{\belowcaptionskip}{-0.2cm}
\begin{center}
	\includegraphics[width=1\linewidth]{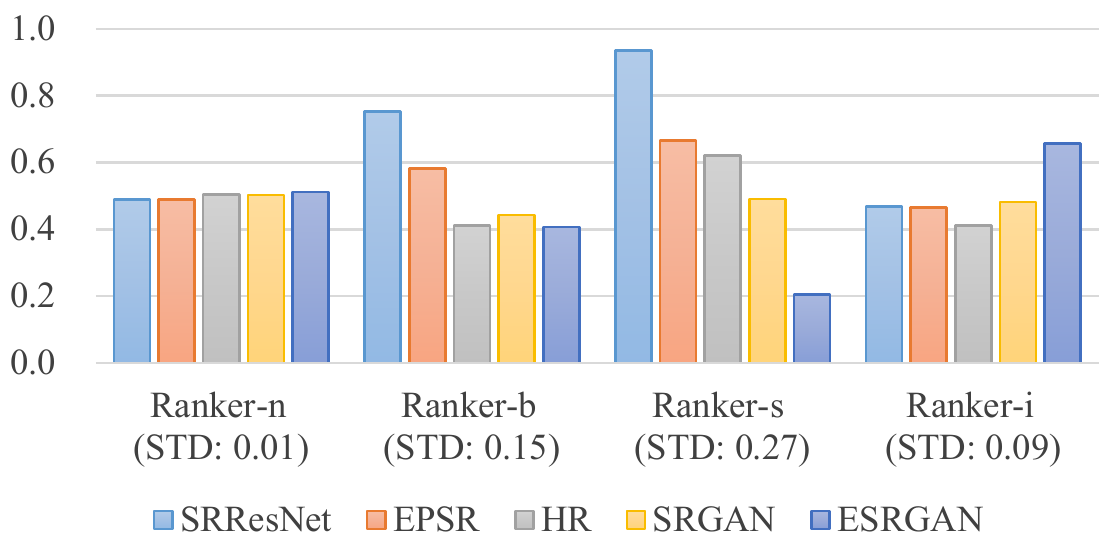} 
\end{center}
   \caption{The normalized distributions of ranking score in Rank100 dataset. [Ranker-n, Ranker-b, Ranker-s, Ranker-i]: Ranker trained by rank dataset with [noise distortion, blur distortion, image SR and image interpolation]. }
\label{fig:rank100}

\end{figure}

\subsection{Evaluation of General Rank Datasets}

\label{section: General Rank}
In this subsection, we evaluate different kinds of rank datasets proposed in Section 4. Specifically, there are three types of datasets using image interpolation, Gaussian noise and Gaussian blur. The Ranker/RankSRGAN trained using these datasets are denoted as Ranker-i/RankSRGAN-Ri, Ranker-n/RankSRGAN-Rn, Ranker-b/RankSRGAN-Rb, respectively. To facilitate comparison, the standard Ranker/RankSRGAN is renamed as Ranker-s/RankSRGAN-Rs. We will first test the ranking ability of different Rankers, and then compare the performance of different RankSRGANs.

\textbf{Ranker.}
As these three rank datasets do not rely on specific perceptual metrics, it is hard to evaluate the trained Rankers only using ground truth labels. In fact, all Rankers can output the right ranking labels with high accuracy. To overcome this difficulty, we propose a validation dataset -- Rank100 to test ``whether the Rankers trained without SR algorithms can classify different SR results''. The proposed Rank100 consists of 20 reference images selected from the DIV2K validation dataset. Each reference image is further processed using bicubic downsampling and $\times$4 SR algorithms, namely SRResNet, SRGAN, ESRGAN and EPSR \cite{vasu2018analyzing}. 

Figure \ref{fig:rank100} shows the average scores of Rankers on different valuation images. Obviously, Ranker-s performs best, as it uses SR algorithms in the rank dataset. The five columns of Rank-s are easily distinguishable from each other, with a high standard deviation -- 0.27. This can be seen as the upper bound. In contrast, the ranking scores of Ranker-n have almost no difference in Rank100. Ranker-b could distinguish SRResNet from EPSR, but has difficulty in classifying SRGAN and ESRGAN. The results of Ranker-i shows that HR images  can get the best score compared with SRGAN and ESRGAN output.   


\textbf{RankSRGAN.} Then we test the performance of RankSRGAN with different rank datasets. The convergence curves of four RankSRGANs are shown in Figure \ref{fig:RankSRGANIQA}. It is evident that all RankSRGANs could outperform SRGAN, which demonstrates the effectiveness of Rankers. Among these models, RankSRGAN-Rs and RankSRGAN-Ri achieve the best performance, while RankSRGAN-Rb is the most stable one. In comparison, RankSRGAN-Rn is only marginally better than SRGAN. This can be attributed to the behaviour of Ranker-n, which cannot distinguish different SR algorithms. The qualitative results of these RankSRGANs are shown in Figure \ref{fig: evaluation2}. We could observe that the outputs of RankSRGAN-Rs and RankSRGAN-Ri are perceptually similar. RankSRGAN-Rb produces images with sharp edges and smooth transitions, but it is inferior in complex details. We can select the desired output image according to their characteristics. 

%
%


\begin{figure*}[htbp]
\begin{center}

\subfigure[Combing different metrics numerically. ]{\includegraphics[width=0.53\linewidth]{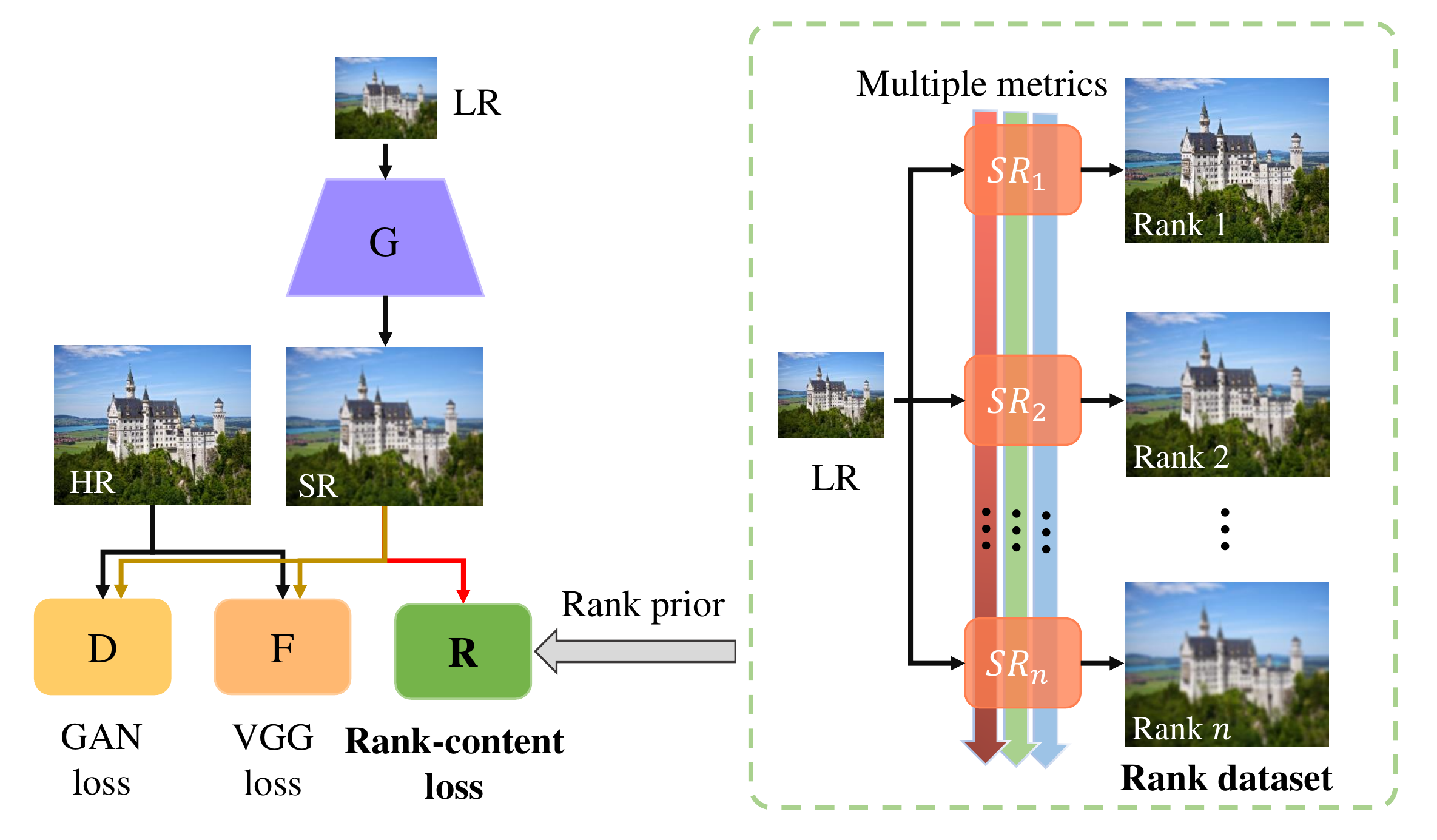}}
\quad
\subfigure[The multiple Rankers with different rank datasets for generator optimization in multiple dimensions. ]{\includegraphics[width=0.43\linewidth]{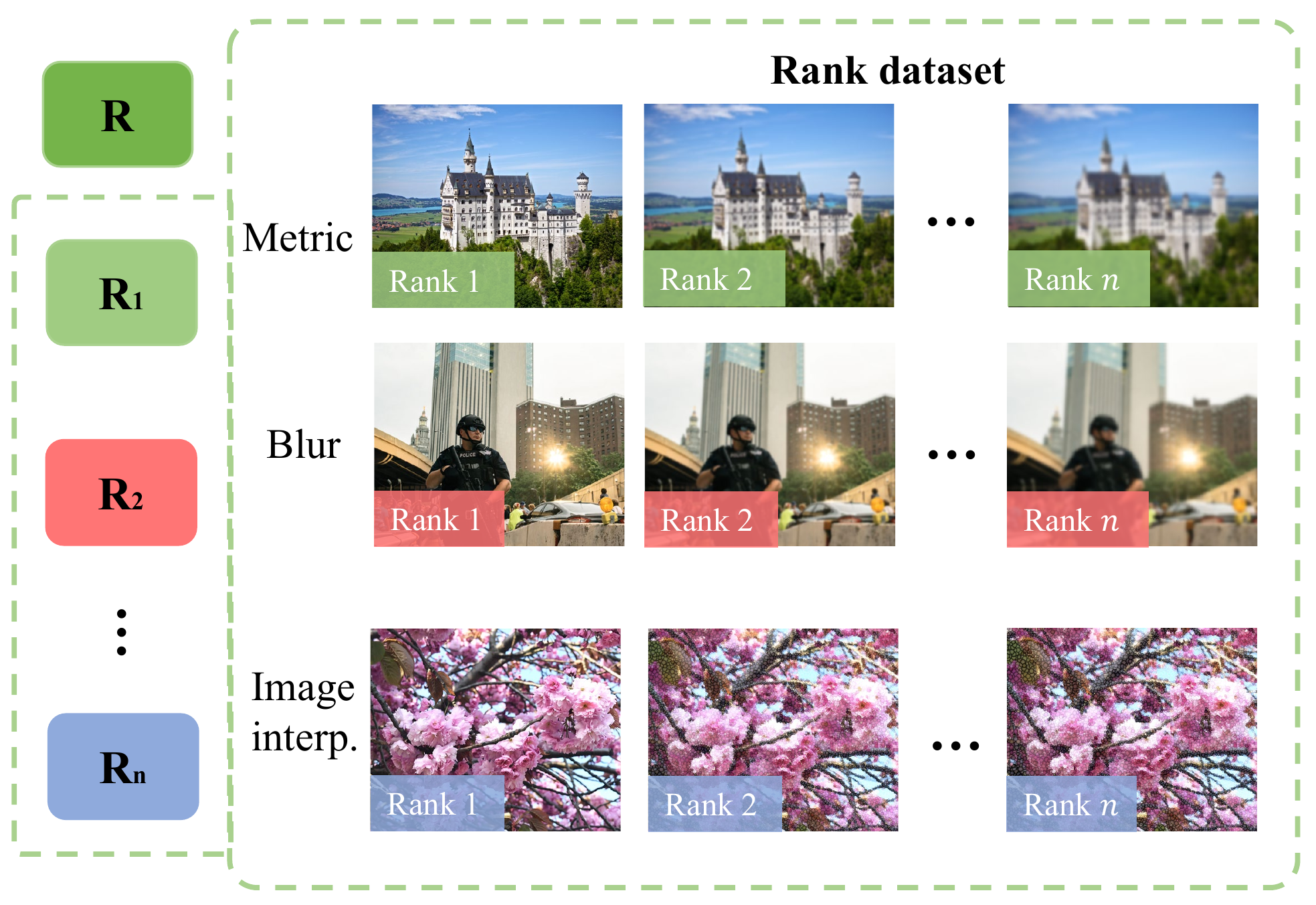}}

\end{center}
\caption{RankSRGAN can be extended to multi-dimensional optimization for different metrics.}
\label{fig:ranksrganmultiple}
\end{figure*}

\begin{figure*}[h!]
\centering

	\includegraphics[width=0.92\linewidth]{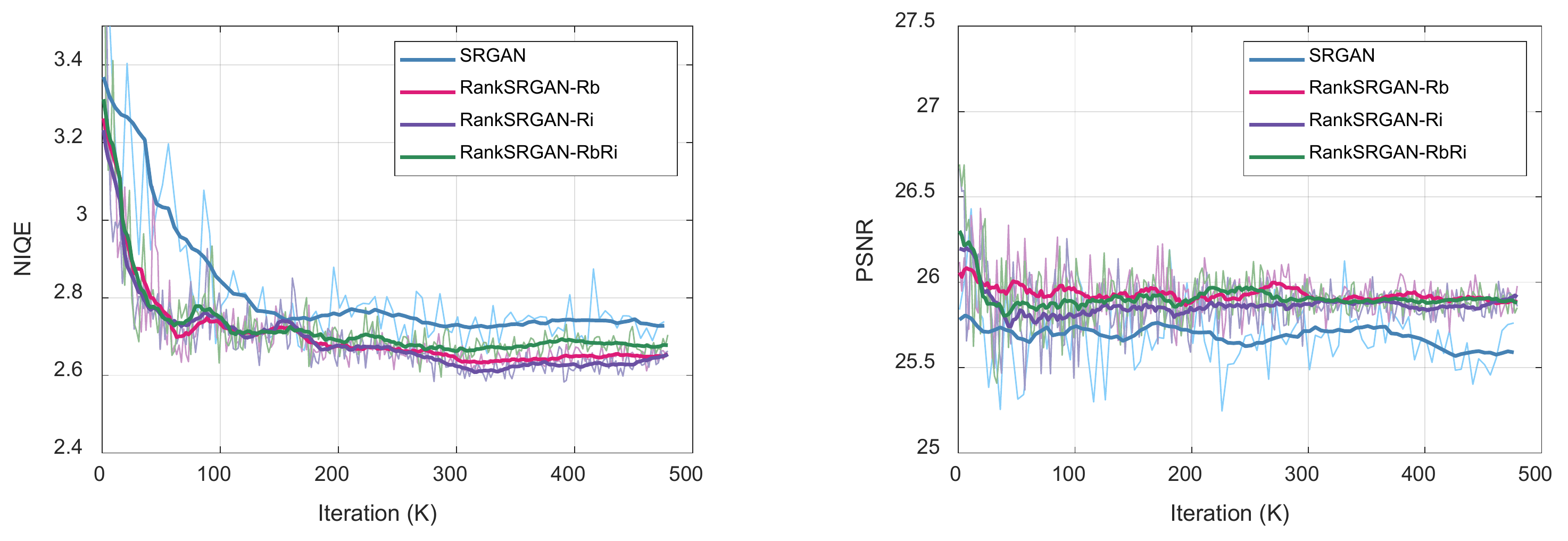} 

\caption{ The convergence curves of SRGAN, RankSRGAN-Rn, RankSRGAN-Rb, RankSRGAN-Ri and RankSRGAN-RbRi. }
\label{fig:RankSRGANRt}
\end{figure*}

\section{RankSRGAN in Multiple Dimensions} 
In the above sections, we have described how to use the proposed Ranker to influence the optimization direction of SRGAN. With a determined perceptual metric and rank dataset, RankSRGAN could produce images in the desired orientation. For example, using NIQE as the perceptual metric will lead to better NIQE scores. Nevertheless, the improvement of NIQE may also affect the performance of other metrics, such as MSE. Furthermore, applying different perceptual constraints will lead to different visual quality, as shown in Figure \ref{fig: evaluation2}. Considering the above dilemma, we make a step further and extend our method to realize multi-dimensional control. It indicates that the model can simultaneously favour different subjective metrics (e.g., MSE and NIQE), and generate images with combined imagery effects (e.g., more details and less artifacts). {We extend our Ranker method to multiple dimensions by adopting combinations of multiple Rankers, where each Ranker is for one metric. Such strategy is simple yet effective. Specifically, we apply multiple Rankers trained on different rank datasets and integrate them in one framework. Note that, in previous sections, we also claim that simply combining different metrics numerically is also effective for achieve various performance. However, by introducing multiple Rankers, We can realize more accurate multi-dimensional control and achieve better results.} The proposed multi-dimensional learning scheme is depicted in Figure \ref{fig:ranksrganmultiple}.

\begin{figure*}[h!]
\begin{center}
	\includegraphics[width=0.9\linewidth]{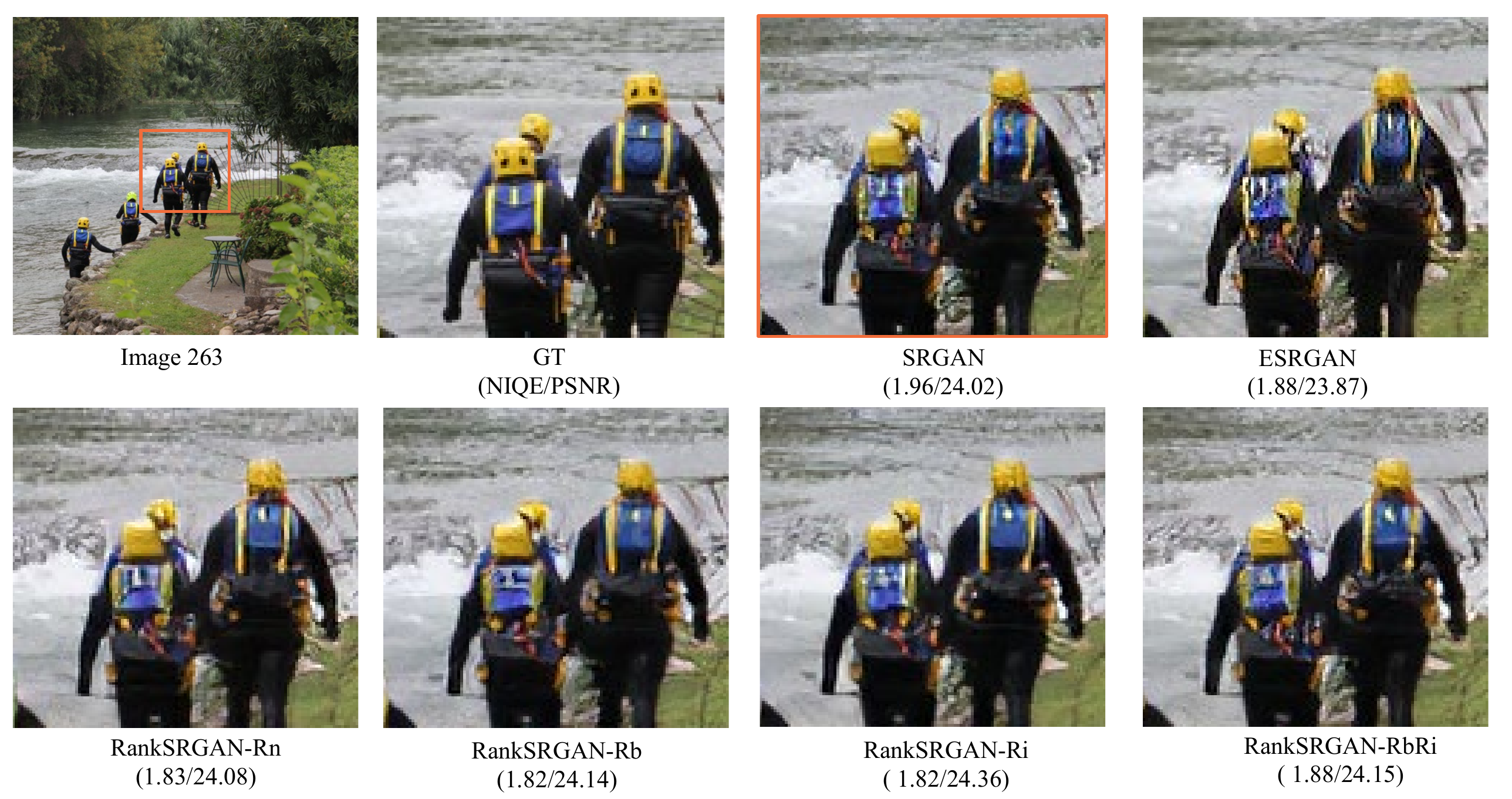} 
\end{center}
\caption{ The visual results of SRGAN, RankSRGAN-Rn, RankSRGAN-Rb, RankSRGAN-Ri and RankSRGAN-RbRi. }
\label{fig:RanktoolVS2}
\end{figure*}

\begin{figure*}[h!]
\begin{center}
	\includegraphics[width=0.9\linewidth]{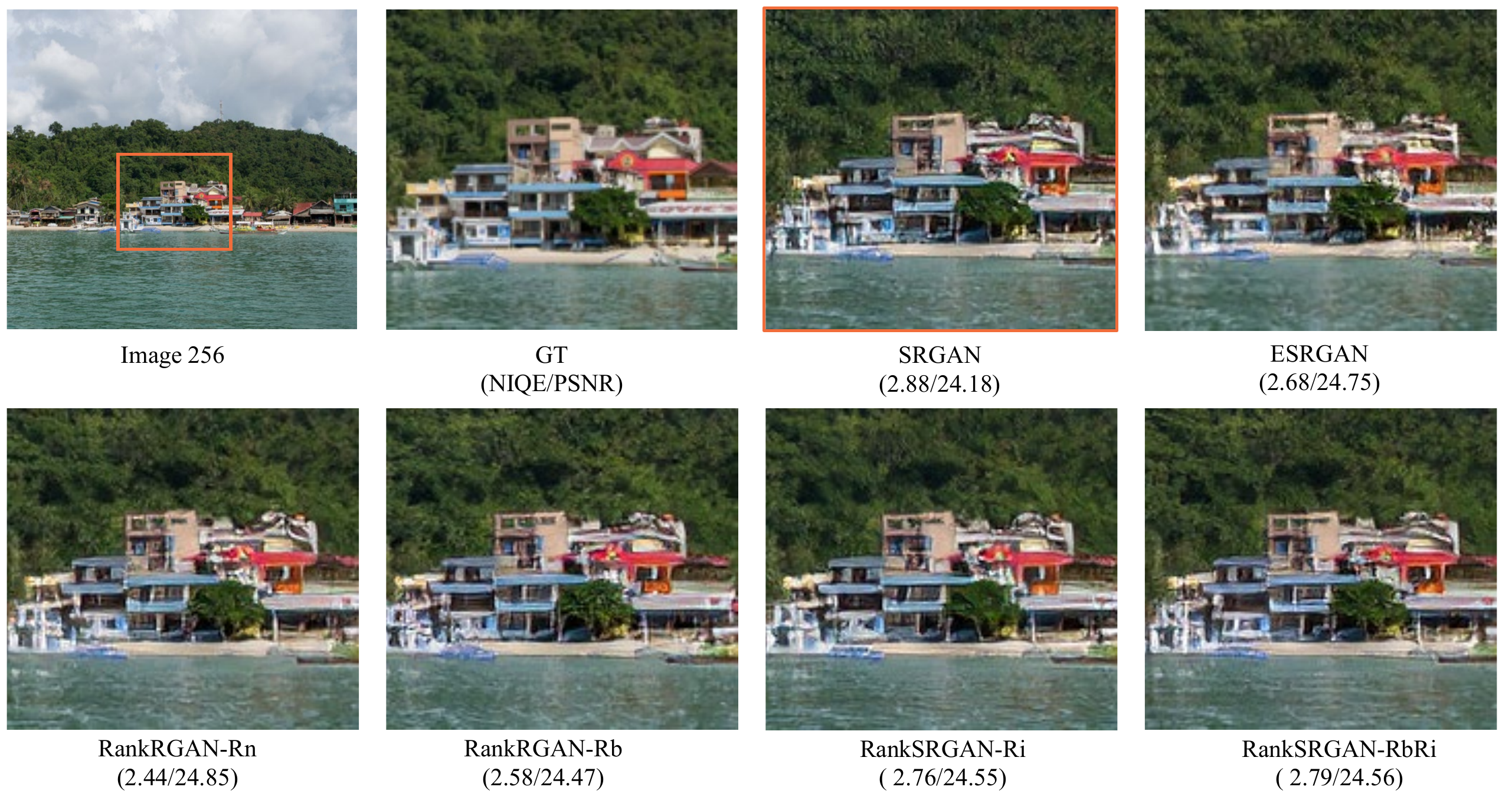} 
\end{center}
\caption{ The visual results of SRGAN, RankSRGAN-Rn, RankSRGAN-Rb, RankSRGAN-Ri and RankSRGAN-RbRi. }
\label{fig:RanktoolVS1}
\end{figure*}



\textbf{Multiple Rankers.}
\label{Multiple Rankers}
We can employ multiple Rankers to constrain the SRGAN model. It is easy to generate a {series} of independent Rankers using different rank datasets, as described in Section \ref{section:rankdataset}. These Rankers have different characteristics, such as favouring sharp edges, less artifacts or smooth transitions. It is flexible to add, delete and reweight these Rankers in SRGAN to achieve different effects. 

To validate the above comment, we select two Rankers: one with Gaussian blur, and the other with image interpolation. We apply both Rankers in the SRGAN model with equal importance. Figure \ref{fig:RankSRGANRt} shows the convergence curves of different RankSRGAN with Gaussian blur (RankSRGAN-Rb), image interpolation (RankSRGAN-Ri) and their combination (RankSRGAN-RbRi). Compared with the baseline SRGAN, all these three RankSRGANs achieve stable improvement in NIQE and PSNR. As the above rank datasets do NOT depend on perceptual metrics (e.g., NIQE), it is better to compare them on visual quality, as in Figure \ref{fig:RanktoolVS2} and \ref{fig:RanktoolVS1}. Obviously, RankSRGAN-Rb could generate sharp edges and favour the image smoothness. RankSRGAN-Ri tends to produce less GAN artifacts, which is influenced by image interpolation. RankSRGAN-RbRi could combine these two effects and generate more realistic images. Specifically, the red strap on the knapsack is closer to the ground truth image, and the overall image has less GAN artifacts. In Figure \ref{fig:RanktoolVS1}, the edges of eaves are closer to the ground truth image, and the textures of {trees} are more natural.

\section{Conclusion}
For perceptual super-resolution, we propose RankSRGAN to optimize SR model in the orientation of perceptual metrics. The key idea is introducing a Ranker to learn the behaviour of the perceptual metrics by learning to rank approach. Moreover, our proposed method can combine the strengths of different SR methods and generate better results. We show detailed analysis on the working mechanism of the proposed Ranker.  To further take advantage of different Rankers, we have also introduced multiple Rankers to guide generator to be optimized towards various metrics. Extensive experiments well demonstrate that our RankSRGAN is a flexible framework, which can achieve superior performance over state-of-the-art methods in optimizing perceptual metrics and have the ability to recover more visually pleasing results with realistic textures. 

\ifCLASSOPTIONcompsoc
  \section*{Acknowledgments}
\else
  \section*{Acknowledgment}
\fi

This work is partially supported by National Natural Science Foundation of China (61906184), the Science and Technology Service Network Initiative of Chinese Academy of Sciences (KFJ-STS-QYZX-092), the Shanghai Committee of Science and Technology, China (Grant No. 21DZ1100100).

\ifCLASSOPTIONcaptionsoff
  \newpage
\fi



\bibliographystyle{IEEEtran}
\bibliography{IEEEabrv,RankSRGAN}

%

%

\begin{IEEEbiography}[{\includegraphics[width=1in,height=1.5in,clip,keepaspectratio]{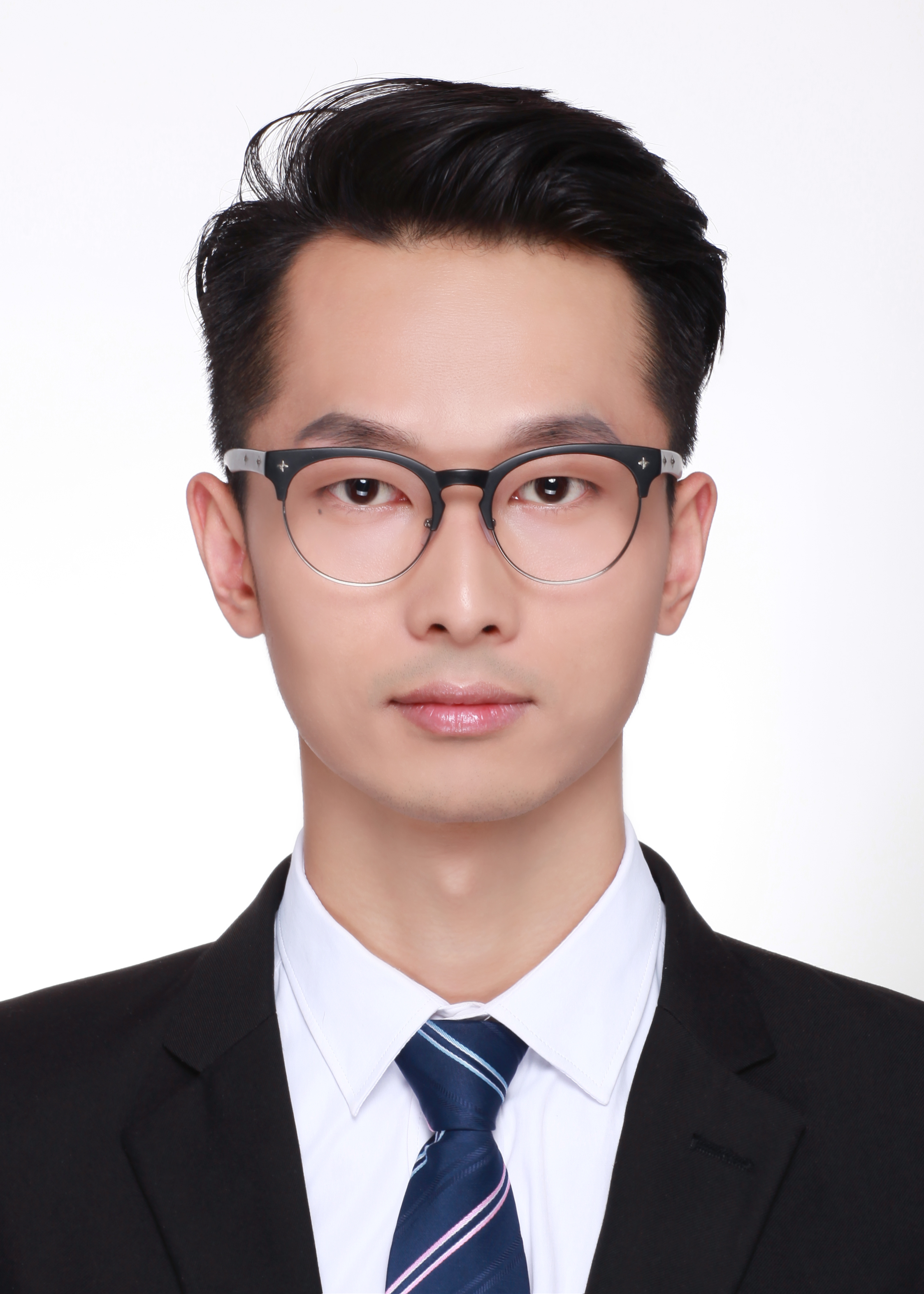}}]{Wenlong Zhang} is currently a Ph.D student at HongKong Polytechnic University.
He received his the B.E. degree from the Zhengzhou University of Light Industry, Zhengzhou, in 2016, and M.S. degree from Beijing Institute of Technology, Beijing, in 2018. He worked as research assistant in Multimedia Laboratory, Shenzhen Institute of Advanced Technology, Chinese Academy of Sciences from 2018 to 2020. His research interests include image super-resolution and restoration.\end{IEEEbiography}

\begin{IEEEbiography}[{\includegraphics[width=1in,height=1.5in,clip,keepaspectratio]{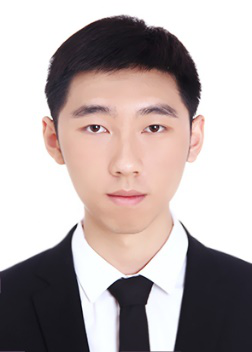}}]{Yihao Liu}
 received his B.E. degree from the University of Chinese Academy of Sciences, Beijing, in 2018. He is now working toward the Ph.D. degree in Multimedia Laboratory, Shenzhen Institute of Advanced Technology, Chinese Academy of Sciences. His research interests include image/video enhancement, super-resolution, interpolation, dehazing, etc.\end{IEEEbiography}

\begin{IEEEbiography}[{\includegraphics[width=1in,height=1.5in,clip,keepaspectratio]{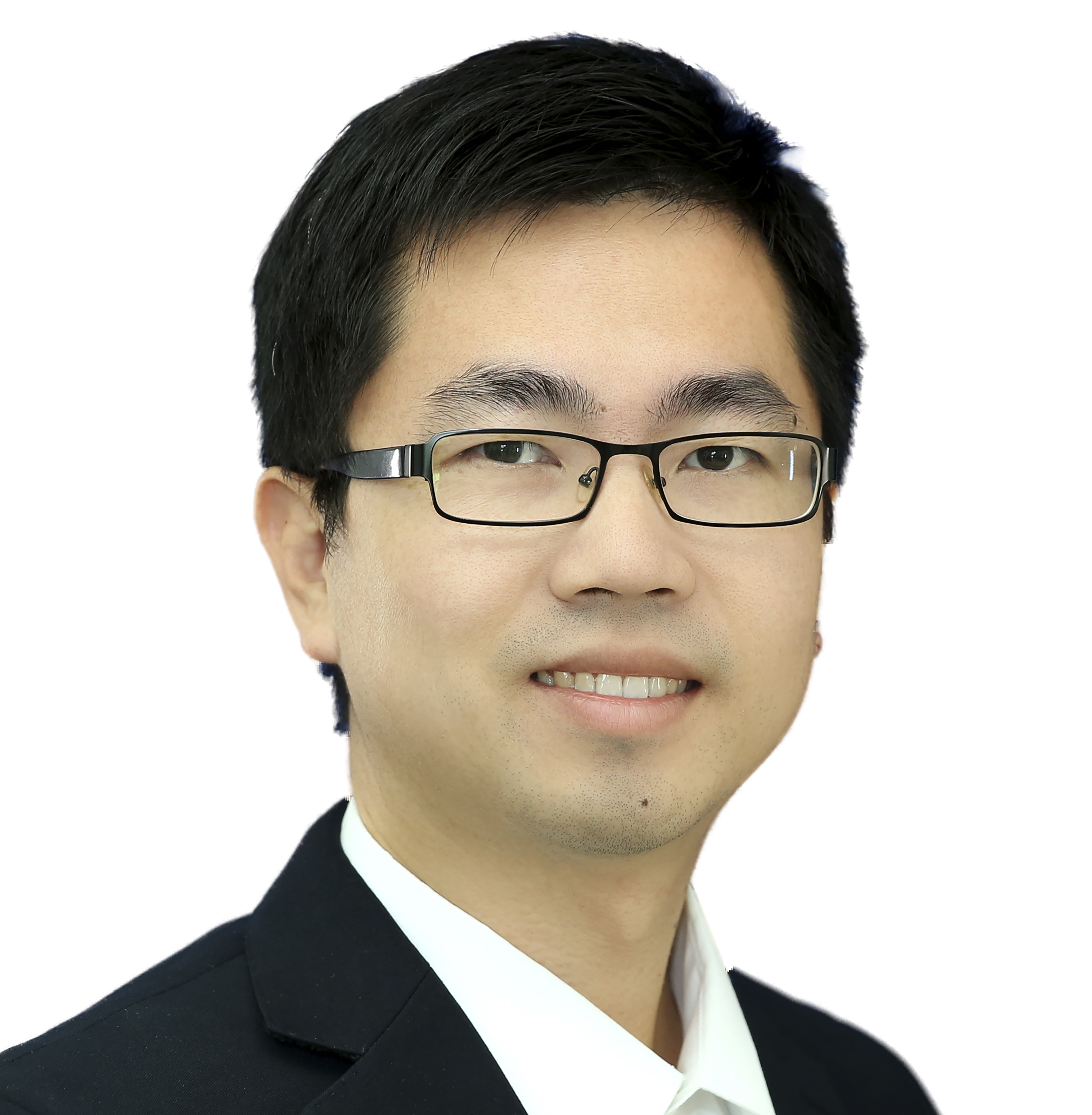}}]{Chao Dong}
 Chao Dong is currently an associate professor in Shenzhen Institute of Advanced Technology, Chinese Academy of Science. He received his Ph.D. degree from The Chinese University of Hong Kong in 2016. In 2014, he first introduced deep learning method -- SRCNN into the super-resolution field. This seminal work was chosen as one of the top ten “Most Popular Articles” of TPAMI in 2016. His team has won several championships in international challenges –NTIRE2018, PIRM2018, NTIRE2019, NTIRE2020 and AIM2020. He worked in SenseTime from 2016 to 2018, as the team leader of Super-Resolution Group. His current research interest focuses on low-level vision problems, such as image/video super-resolution, denoising and enhancement.\end{IEEEbiography}

\begin{IEEEbiography}[{\includegraphics[width=1in,height=1.5in,clip,keepaspectratio]{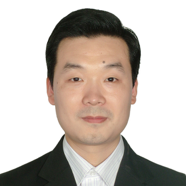}}]{Yu Qiao}
 (Senior Member, IEEE) received the Ph.D. degree from the University of Electro-Communications, Japan, in 2006. He was a JSPS Fellow and a Project Assistant Professor with The University of Tokyo from 2007 to 2010. He is currently a Professor with the Shenzhen Institutes of Advanced Technology, Chinese Academy of Sciences. He has authored over 180 articles in journals and conferences, including PAMI, IJCV, TIP, ICCV, CVPR, ECCV, and AAAI. His research interests include computer vision, deep learning, and intelligent robots. He was a recipient of the Lu Jiaxi Young Researcher Award from the Chinese Academy of Sciences in 2012. He was the winner of video classification task in the ActivityNet Large Scale Activity Recognition Challenge 2016 and the first runner-up of scene recognition task in the ImageNet Large Scale Visual Recognition Challenge 2015.
\end{IEEEbiography}






\end{document}